\documentclass[10pt,letterpaper]{article}
\usepackage{harnessvln_preprint}

\usepackage{amsmath,amsfonts,bm}

\def\eqref#1{equation~\ref{#1}}

\def\1{\bm{1}}

\DeclareMathAlphabet{\mathsfit}{\encodingdefault}{\sfdefault}{m}{sl}
\SetMathAlphabet{\mathsfit}{bold}{\encodingdefault}{\sfdefault}{bx}{n}

\usepackage{algorithm}
\usepackage{algpseudocode}
\usepackage{booktabs,multirow,graphicx,tabularx,array}
\usepackage{pifont}
\definecolor{tablegray}{HTML}{707070}
\newcommand{\cmark}{\ding{51}}
\newcommand{\xmark}{\ding{55}}

\usepackage{listings}
\usepackage{tcolorbox}
\tcbuselibrary{breakable,skins}

\newtcolorbox{appPrompt}[1]{
  enhanced,
  breakable,
  title={#1},
  fonttitle=\small\bfseries,
  fontupper=\small,
  coltitle=black,
  colback=white,
  colbacktitle=black!4,
  colframe=black!25,
  boxrule=0.4pt,
  titlerule=0.3pt,
  arc=1mm,
  boxsep=0pt,
  left=3mm, right=3mm,
  top=2.5mm, bottom=2.5mm,
  toptitle=1.8mm, bottomtitle=1.8mm,
  before upper={\setlength{\parindent}{0pt}},
  before skip=10pt, after skip=10pt
}

\lstdefinestyle{appJSON}{
  basicstyle=\ttfamily\footnotesize,
  columns=fullflexible,
  keepspaces=true,
  showstringspaces=false,
  breaklines=true,
  breakatwhitespace=false,
  backgroundcolor=\color{black!3},
  frame=single,
  framerule=0pt,
  framesep=5pt,
  xleftmargin=5pt,
  xrightmargin=5pt,
  aboveskip=8pt,
  belowskip=2pt
}

\title{HarnessVLN: Unifying Training-Free Embodied Navigation through an Agent Harness}
\author{%
  Yang Chen\textsuperscript{1,2}\quad
  Lirong Che\textsuperscript{2,3}\quad
  Zhenyu Huang\textsuperscript{1}\quad
  Wenbo Fu\textsuperscript{1}\quad
  Chuang Wang\textsuperscript{2}\\[4pt]
  Xu Cao\textsuperscript{2}\quad
  Daqi Liu\textsuperscript{2}\quad
  Yuzhe Yang\textsuperscript{2}\quad
  Jian Su\textsuperscript{2,*}\quad
  Lan-Zhe Guo\textsuperscript{1,*}\\[9pt]
  {\small\textsuperscript{1}Nanjing University\quad
  \textsuperscript{2}AGIBOT\quad
  \textsuperscript{3}Tsinghua University}\\[4pt]
  {\small\textsuperscript{*}Corresponding authors}
}
\date{}
\hypersetup{
  pdftitle={HarnessVLN: Unifying Training-Free Embodied Navigation through an Agent Harness},
  pdfauthor={Yang Chen, Lirong Che, Zhenyu Huang, Wenbo Fu, Chuang Wang, Xu Cao, Daqi Liu, Yuzhe Yang, Jian Su, Lan-Zhe Guo},
  pdfsubject={Training-free embodied navigation},
  pdfkeywords={HarnessVLN, embodied navigation, agent harness, vision-language navigation}
}

\begin{document}

\maketitle

\begin{figure}[htbp]
  \centering
  \includegraphics[width=0.94\linewidth]{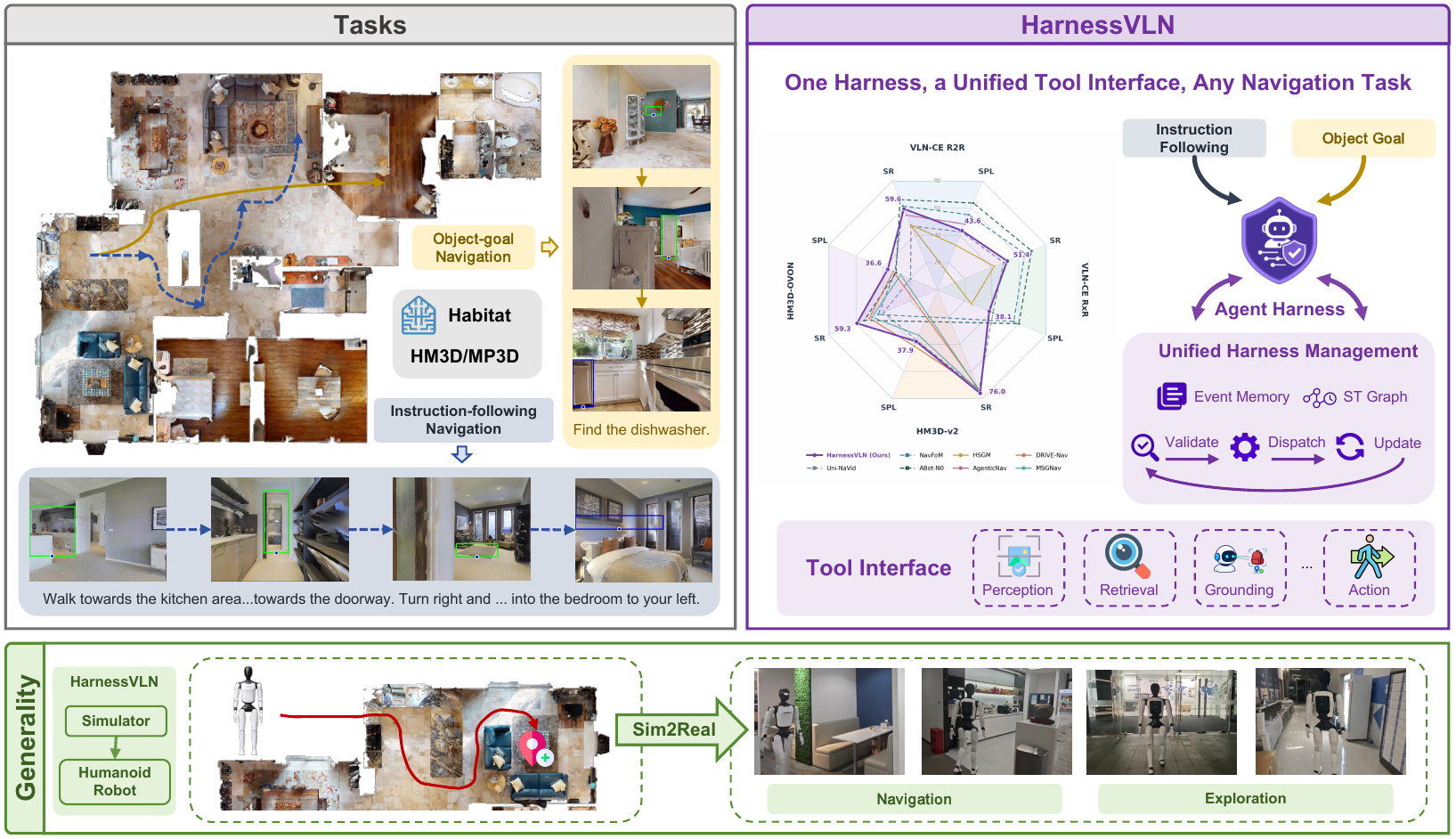}
  \caption{\textbf{HarnessVLN: One Harness for Navigation across Tasks and Environments.} A unified Agent Harness and tool interface support navigation tasks, delivering strong performance across four benchmarks and enabling deployment on a humanoid robot in real-world environments.}
  \label{fig:teaser}
\end{figure}

\begin{abstract}
Embodied navigation requires agents to ground instructions or object goals in spatial observations and translate plans into successful execution. As multimodal large language models (MLLMs) become increasingly capable, they offer stronger support for navigation without task-specific training; however, improved semantic reasoning alone does not ensure that proposed actions remain consistent with spatial evidence, task progress, and execution outcomes. We introduce HarnessVLN, a zero-shot, training-free framework that unifies instruction-following and object-goal navigation through a shared Agent Harness. The Harness coordinates perception, memory, and execution tools through a unified interface, validating planner proposals for evidential support, geometric feasibility, and subgoal consistency before dispatch. It jointly manages hierarchical event memory and a persistent Spatiotemporal Graph to track task progress, preserve spatial evidence, and contextualize failures. Structured execution feedback updates this shared state, guiding subsequent planning, recovery, and termination. Across R2R, RxR, HM3D-v2, and HM3D-OVON, HarnessVLN achieves success rates of 59.6\%, 51.4\%, 76.0\%, and 59.3\%, respectively, outperforming prior training-free methods. Humanoid robot deployment further demonstrates its applicability to both navigation tasks in real-world environments. The project page is available at \url{https://agibot-harnessvln.netlify.app/}.
\end{abstract}

\section{Introduction}
\label{sec:introduction}

Navigation is a foundational capability of embodied agents, enabling them to reach specified locations for further interaction with the physical world. Instruction-following and object-goal navigation represent two central forms of this capability, requiring agents to follow a route described in language or locate a specified object, respectively~\citep{uninavid,abot}. Despite their different task definitions, both require agents to ground language in partially observed environments, continuously accumulate spatial knowledge, and ultimately reach the destination.

Training-based methods acquire these capabilities from task-specific demonstrations or large-scale trajectory data, mapping visual observations and language inputs to navigation actions~\citep{etpnav,mtu3d,ground,lookstep,filmnav}. Although effective on specific tasks, extending these methods to new task formulations or environments requires additional training data for adaptation. Pretrained multimodal large language models (MLLMs) offer an alternative: leveraging their semantic knowledge and reasoning capabilities to interpret instructions, identify targets, and guide exploration without task-specific navigation training~\citep{vlfm,tango,hsgm,stegnav,msgnav}.

However, in existing MLLM-based navigation systems, the model primarily serves as a planner within a task pipeline: perception modules provide observations, the planner proposes subgoals, and downstream modules perform spatial grounding and action execution. This design relies heavily on the MLLM's semantic reasoning, while evidence retention, proposal validation, and failure handling lack a unified mechanism for coordinating their responsibilities. This exposes a central challenge: ensuring that the actions proposed by the MLLM advance the agent toward task completion as spatial knowledge and navigation progress continually evolve.

This challenge becomes more pronounced as navigation progresses: execution failures can lead to discrepancies between planned behavior and the actual task state. Moreover, recognizing a target does not confirm arrival, and issuing an action does not establish completion of the corresponding subgoal. Addressing these gaps requires a runtime mechanism that jointly tracks spatial evidence, task progress, and execution outcomes. Agent harnesses provide a foundation for such coordination by integrating reasoning models, tools, memory, and environmental feedback within a unified framework~\citep{aspire,harnessvla,embodiedagent}. Adapting this framework to navigation requires explicit spatial and temporal grounding, with evidence retaining information about where and when it was acquired. Proposed targets must be reachable and relevant to the task, while execution outcomes must guide subsequent planning, recovery, and termination decisions.

We introduce \textbf{HarnessVLN}, a zero-shot, training-free framework that supports instruction-following and object-goal navigation through a shared \emph{Agent Harness}, as illustrated in Figure~\ref{fig:teaser}. The Harness coordinates perception, retrieval, grounding, navigation, recovery, and termination through a unified tool interface. The MLLM planner proposes semantic operations, which the Harness validates against supporting evidence, geometric feasibility, the active subgoal, and relevant failure history before dispatch. Structured execution feedback then updates the agent's state and informs the next decision. This shared interaction protocol accommodates both established navigation tasks through task-specific progress representations and completion criteria.

HarnessVLN maintains two complementary forms of memory to sustain coordination across decisions. Hierarchical event memory records decision context, subgoal progress, and execution events, while a persistent Spatiotemporal Graph (ST Graph) organizes spatial evidence, observation provenance, and associated failure records. Together, they enable the Harness to retrieve relevant experience, validate proposals against accumulated evidence, and guide recovery after failed attempts. A replaceable Navigation Executor converts validated targets into paths and motion commands, while stop validation checks whether the available semantic and spatial evidence supports task completion.

We evaluate HarnessVLN on VLN-CE R2R and RxR for instruction-following navigation and on HM3D-v2 and HM3D-OVON for object-goal navigation~\citep{r2r,rxr,hm3dsem,hm3dovon}. Using the same framework across four benchmarks, HarnessVLN achieves success rates of 59.6\%, 51.4\%, 76.0\%, and 59.3\%, respectively, demonstrating its effectiveness across instruction-following and object-goal navigation without task-specific training. Deployment on a humanoid robot demonstrates its applicability to real-world navigation.

Our main contributions are:
\begin{itemize}
\item We introduce HarnessVLN, a training-free framework that supports instruction-following and object-goal navigation through a shared Agent Harness and a unified tool interface.
\item We develop a navigation runtime that combines hierarchical event memory and a persistent ST Graph with proposal validation and execution feedback, supporting evidence-grounded action, failure recovery, and verified termination.
\item We demonstrate improved performance over prior training-free methods across four navigation benchmarks and validate the framework's real-world applicability through humanoid robot deployment.
\end{itemize}

\section{Related Work}
\label{sec:related_work}

\paragraph{Training-Based Embodied Navigation.}
Training-based methods learn navigation policies from task-specific demonstrations or large-scale trajectory data~\citep{mapdream,streamvln,navila,janusvln,omninav}. NaVid~\citep{navid} predicts actions from monocular video and instructions, while Uni-NaVid~\citep{uninavid} and NavFoM~\citep{navfom} extend unified policy learning across tasks and platforms. Although effective, these approaches rely on navigation-specific data and training, and adapting them to new task settings may require additional supervision. HarnessVLN requires no task-specific navigation training, using a general-purpose MLLM for semantic planning and a replaceable Navigation Executor for path planning and control.

\paragraph{Zero-Shot Embodied Navigation.}
Zero-shot navigation transfers semantic knowledge from pretrained language and vision models without task-specific navigation training. Object-goal methods combine open-vocabulary perception or commonsense reasoning with mapping and exploration~\citep{cow,esc,vlfm,apexnav}, while instruction-following methods use language models for instruction decomposition, progress tracking, and target selection~\citep{instructnav,canav,lavira,drivenav,smartway,evonav}. 

Agentic embodied systems coordinate pretrained models, tools, memory, and execution feedback beyond monolithic policies~\citep{codeaspolicies,voxposer,saycan,cap-x,roboclaw,agenticrobot,re2agent,lifting,embodiedagent}.  AgenticNav~\citep{agenticnav} exposes action, depth, and selective visual recall as callable tools with geometric safety checks, while Uni-LaViRA~\citep{unilavira} unifies navigation tasks through hierarchical action translation, TODO List Memory, and Second Chance Backtrack. HarnessVLN unifies tools, task state, and structured memory through feedback-driven updates to task progress and event memory, with a persistent ST Graph linking observations and failures for cross-subgoal retrieval. A shared runtime validates actions and stopping requests against spatial evidence and task progress in both instruction-following and object-goal navigation.

\begin{figure}[htbp]
  \centering
  \includegraphics[width=\linewidth]{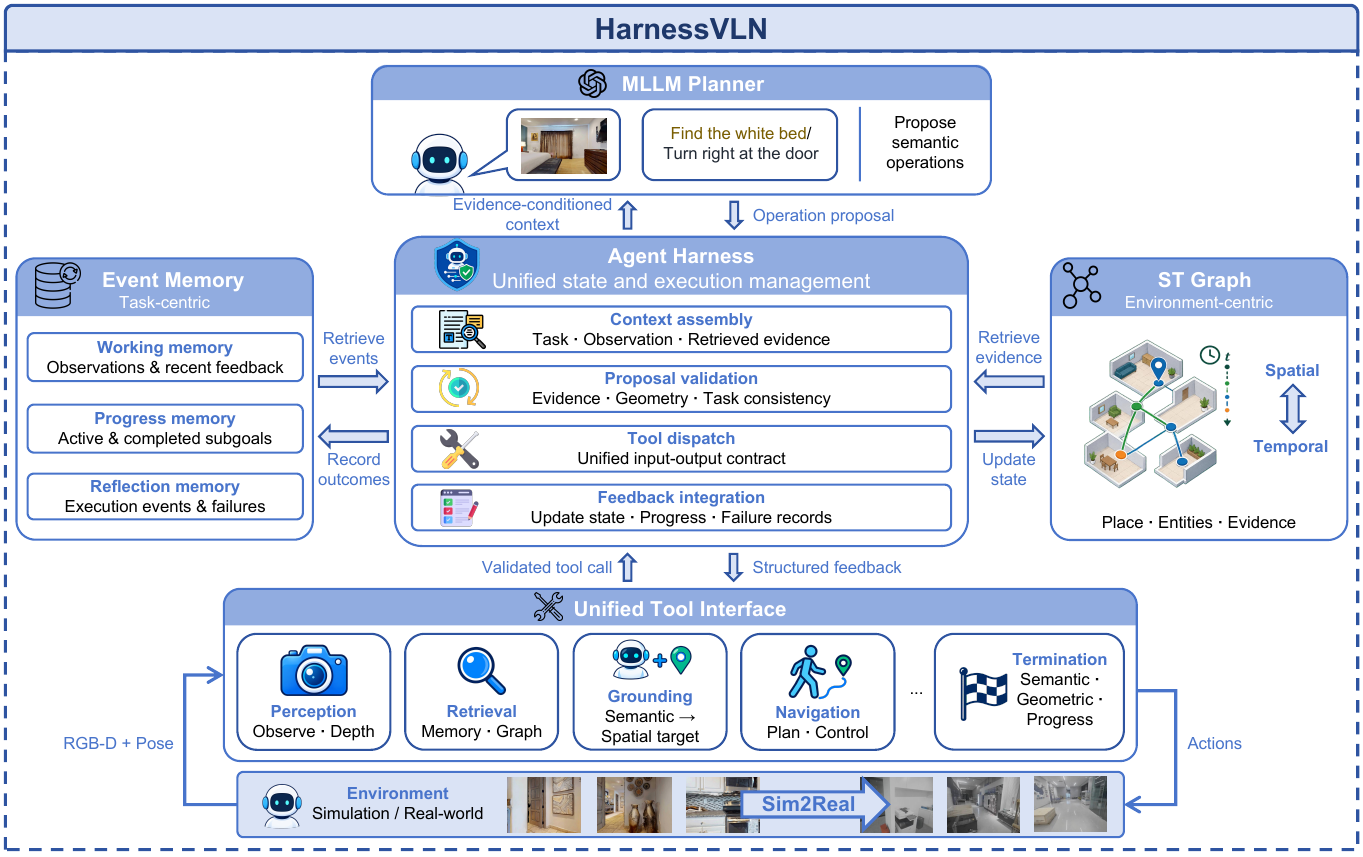}
  \caption{Overview of HarnessVLN. The Agent Harness coordinates MLLM planning, hierarchical event memory, and the ST Graph through a unified tool interface, forming a closed loop of context assembly, proposal validation, tool execution, and feedback integration.}
  \label{fig:framework}
\end{figure}

\section{HarnessVLN: A Harness-Mediated Embodied Navigation Agent}
\label{sec:method}

As shown in Fig.~\ref{fig:framework}, HarnessVLN uses an \emph{Agent Harness} to coordinate planning, memory, and tool execution. We introduce its decision process, tool interface, and event memory (Section~\ref{sec:agent_harness}), followed by the Harness-managed ST Graph for spatial retrieval and updates (Section~\ref{sec:spatiotemporal_graph}), and grounded execution with evidence-based termination (Section~\ref{sec:closed_loop_navigation}).

\subsection{Agent Harness}
\label{sec:agent_harness}

\subsubsection{Harness-Mediated Decision Making}
\label{sec:harness_decision}

At decision step $t$, the Harness maintains the state
\begin{equation}
    \mathcal{S}_t =
    \bigl(x,\, \omega_t,\, p_t,\, z_t,\, \mathcal{M}_t,\, \mathcal{G}_t\bigr),
    \label{eq:harness_state}
\end{equation}
where $x$ denotes the task specification, namely a route instruction or target object category; $\omega_t$ is the current pose-aligned RGB-D observation and local geometric map; $p_t$ is the agent pose; $z_t$ represents task progress; $\mathcal{M}_t$ is a hierarchical event memory that records task-centric events and their execution contexts; and $\mathcal{G}_t$ is a persistent ST Graph that organizes environment-centric relations and evidence.

HarnessVLN provides a unified information interface for instruction following and ObjectNav, yielding an evidence-conditioned decision cycle. For instruction following, $z_t$ records the active route segment, satisfied motion constraints, and observed landmarks. For ObjectNav, it tracks explored regions, candidate target hypotheses, and their verification status. These task-specific variables affect planning and termination criteria without changing the underlying Harness protocol.

\subsubsection{Unified Tool Interface}

As summarized in Table~\ref{tab:harness_tools}, HarnessVLN exposes heterogeneous navigation capabilities through a unified tool interface. Perception tools collect directional or panoramic RGB-D observations; retrieval and grounding tools obtain task-relevant evidence from the current observation and the ST Graph and ground semantic subgoals to executable targets; navigation and recovery tools invoke the Navigation Executor or return the agent to a previously visited location; and the termination tool verifies task completion before stopping. This shared input--output contract allows individual tools to be replaced without modifying the MLLM Planner or the Harness protocol.

Before dispatching a tool call, the Harness validates its arguments, the provenance and recency of its supporting evidence, the geometric feasibility of the proposed target, and its consistency with the active subgoal. Each tool returns structured feedback containing the execution status, relevant measurements, and failure evidence. The Harness uses this feedback to update its internal state and the ST Graph, providing an updated context for the next planning decision.

\begin{table}[t]
    \centering
    \caption{Core tools exposed by HarnessVLN through a unified interface.}
    \label{tab:harness_tools}
    \small
    \setlength{\tabcolsep}{4pt}
    \renewcommand{\arraystretch}{1.10}
    \begin{tabularx}{\linewidth}{
        @{}
        >{\raggedright\arraybackslash}p{0.27\linewidth}
        >{\raggedright\arraybackslash}p{0.17\linewidth}
        >{\raggedright\arraybackslash}X
        @{}
    }
        \toprule
        \textbf{Tool} & \textbf{Role} & \textbf{Function} \\
        \midrule

        \texttt{observe}
        & Perception
        & Capture the current RGB-D observation and agent pose. \\

        \texttt{observe\_panorama}
        & Perception
        & Capture panoramic observations at predefined headings. \\

        \texttt{retrieve\_memory}
        & Retrieval
        & Retrieve task-relevant evidence from memory and graph. \\

        \texttt{ground\_target}
        & Grounding
        & Ground a subgoal to a candidate view and image region. \\

        \texttt{query\_depth}
        & Perception
        & Estimate the depth and uncertainty of a grounded region. \\

        \texttt{navigate\_to}
        & Navigation
        & Invoke the Navigation Executor for a validated target. \\

        \texttt{backtrack}
        & Recovery
        & Return to a previously visited location and verify arrival. \\

        \texttt{request\_stop}
        & Termination
        & Validate task completion before terminating execution. \\

        \bottomrule
    \end{tabularx}
\end{table}

\subsubsection{Hierarchical Event Memory}
\label{sec:hierarchical_memory}

The hierarchical event memory $\mathcal{M}_t$ is a task-centric record of the agent's interaction history. It stores the events required to interpret the current decision, track subgoal completion, and diagnose recent execution outcomes.

\paragraph{Working memory.}
Working memory stores the current observation, a bounded history of panoramic views, the active target hypothesis, and recent tool feedback. It retains only the short-term context needed for the next decision and thus does not grow unboundedly with trajectory length.

\paragraph{Progress memory.}
Progress memory represents task decomposition and completion. Each subgoal has a unique identifier and one of four states: pending, active, completed, or blocked. At most one subgoal is active at a time, and completion must be supported by a pose-aligned observation or a successful tool result. This prevents the planner from advancing the task based on assumptions.

\paragraph{Reflection memory.}
Reflection memory records complete subgoal-specific execution events, including unreachable targets, inconsistent grounding, collisions, lack of progress, failed backtracking, and rejected stopping requests. Each record retains an event identifier, task context, location, timestamp, tool feedback, and supporting observation. Reflection memory remains the authoritative source for these event-level traces. When a failure has reusable spatial implications, the Harness writes only a lightweight annotation and a reference to the original event into the ST Graph. This separation enables failure-aware retrieval without duplicating the full execution history.

\subsection{Harness-Managed Spatiotemporal Graph}
\label{sec:spatiotemporal_graph}

The ST Graph $\mathcal{G}_t$ maintains an environment-centric abstraction of the agent's accumulated experience. Unlike event memory, which preserves task execution history, the graph consolidates reusable spatial structure and time-indexed evidence for replanning and recovery. It persists across subgoals while tracking when and where each observation was acquired, allowing the Harness to distinguish current evidence from stale or superseded hypotheses.

\paragraph{Persistent representation.}
\label{sec:persistent_spatiotemporal_representation}
We define the graph as
\begin{equation}
    \mathcal{G}_t =
    \bigl(\mathcal{V}^{P}_t \cup \mathcal{V}^{E}_t,
          \mathcal{E}_t,\mathcal{T}_t\bigr),
    \label{eq:spatiotemporal_graph}
\end{equation}
where $\mathcal{V}^{P}_t$ and $\mathcal{V}^{E}_t$ are place and entity nodes, $\mathcal{E}_t$ contains typed spatial relations, and $\mathcal{T}_t$ stores timestamps and lightweight event annotations. Place nodes summarize visited locations, supporting observations, and traversable waypoints. Spatially compatible observations are merged into an existing place; otherwise, a new node is created. Consecutive places are connected by bidirectional \textsc{NavigableTo} edges, forming a persistent topology for forward navigation and backtracking.

Entity nodes represent ObjectNav targets and instruction-referenced landmarks through semantic aliases, spatial hypotheses, confidence estimates, and supporting views. \textsc{ObservedFrom} edges preserve the viewpoints and times at which an entity was observed, while \textsc{Contains} edges encode coarse place--entity associations. This provenance allows the Harness to retrieve both a target hypothesis and the evidence required to verify it.

\paragraph{Task-conditioned retrieval.}
\label{sec:entity_place_association}
Before each MLLM decision, the Harness ranks place nodes according to the active subgoal $g_t$:
\begin{equation}
    s(v_i \mid g_t)
    = R(v_i,g_t)
    + \lambda_{\mathrm{rec}} C(v_i)
    + \lambda_{\mathrm{sal}} A(v_i)
    - \lambda_{\mathrm{fail}} P(v_i,g_t),
    \label{eq:retrieval_score}
\end{equation}
where $R$, $C$, and $A$ measure semantic relevance, recency, and spatial salience, respectively, and $P$ penalizes applicable prior failures. The top-$K$ places are augmented with nearby nodes and their associated entities, waypoints, and spatial relations. This produces a compact retrieval set $\mathcal{R}_t$, allowing the graph to grow without increasing the MLLM context with trajectory length.

\paragraph{Failure-aware updates.}
\label{sec:failure_aware_modeling}
Failures with reusable spatial implications are attached to the corresponding place, entity, or relation as lightweight annotations. Each annotation records its subgoal condition, failure type, timestamp, applicability weight, and a reference to the complete event-memory record. Its relevance decreases when the subgoal changes or new evidence invalidates the failure, and increases when consistent failures recur. The graph therefore provides a spatial index for recovery, while event memory remains the complete record of task execution.

\begin{algorithm}[t]
\caption{Harness-Mediated Navigation}
\label{alg:harness_navigation}
\begin{algorithmic}[1]
\Require Task $x$ and initial observation $\omega_0$
\State Initialize $\mathcal{S}_0$, event memory $\mathcal{M}_0$, and
Spatiotemporal Graph $\mathcal{G}_0$
\While{$\mathcal{S}_t$ is non-terminal}
    \If{$\omega_t$ is inconsistent with the current pose}
        \State $\omega_t \gets \Call{Observe}{}$
    \EndIf
    \State $\mathcal{M}_t \gets
        \Call{RecordEvent}{\mathcal{M}_t,\omega_t}$
    \State $\mathcal{G}_t \gets
        \Call{ConsolidateGraph}{\mathcal{G}_t,\mathcal{M}_t}$
    \State $g_t \gets \Call{ActiveSubgoal}{z_t}$
    \State $\mathcal{R}_t \gets
        \Call{RetrieveExperience}{\mathcal{G}_t,g_t,p_t}$
    \State $\phi_t \gets
        \Call{AssembleContext}{\mathcal{S}_t,\mathcal{R}_t}$
    \State $c_t \gets \pi_\theta(\phi_t)$
    \State $\widehat{c}_t \gets
        \Call{Validate}{c_t,\mathcal{S}_t}$
    \State $y_t \gets \Call{DispatchTool}{\widehat{c}_t}$
    \State $\mathcal{S}_{t+1} \gets
        \Call{UpdateHarnessState}{\mathcal{S}_t,y_t}$
\EndWhile
\State \Return verified terminal state and trajectory
\end{algorithmic}
\end{algorithm}

\subsection{Grounded Navigation Execution}
\label{sec:closed_loop_navigation}

The MLLM proposes semantic navigation commands, whereas physical execution requires grounded targets and executable controls. HarnessVLN therefore exposes the Navigation Executor as a Harness-managed tool: the Harness grounds and validates each command, while the Executor performs path planning and local control~\citep{navdp}.

\paragraph{Target grounding and execution.}
\label{sec:tool_mediated_execution}
For each command $c_t$, the Harness resolves its target from the current observation and the ST Graph. A command is dispatched only if the target is supported by available evidence, geometrically reachable, and consistent with the active subgoal and applicable failure history. The Navigation Executor converts the validated target into an executable route and returns structured feedback, such as arrival, collision, unreachability, or lack of progress. This feedback updates the Harness state, event memory, and graph. Forward navigation and backtracking therefore follow the same validation--execution--update loop.

\paragraph{Evidence-based termination.}
\label{sec:evidence_based_termination}
Termination is mediated by the Harness. The MLLM may propose stopping but cannot directly issue an environment-level \textsc{Stop} action. A request is accepted only if
\begin{equation}
    F_{\mathrm{stop}}
    = F_{\mathrm{semantic}}
      \land F_{\mathrm{geometric}}
      \land F_{\mathrm{progress}},
    \label{eq:stop_validation}
\end{equation}
where the three terms verify target identity, geometric validity, and task completion. For object-goal navigation, the target must be visually supported and pass the Harness's geometric proximity checks (Appendix~\ref{app:stop_validation}); these checks are distinct from the benchmark's geodesic success criterion. For instruction following, the current evidence must support the final route segment, referenced landmark, and grounded endpoint. A rejected request is recorded in event memory and triggers further observation, target refinement, approach, or backtracking. Stopping is thus determined by embodied evidence rather than planner confidence alone.

Algorithm~\ref{alg:harness_navigation} summarizes the runtime procedure.

\begin{table*}[t!]
\centering
\caption{Comparison with existing supervised and training-free methods on the VLN-CE R2R and RxR val-unseen splits. $^\dagger$ denotes results evaluated on the R2R Rand100 subset.}
\label{tab:instruction_navigation}

\resizebox{\textwidth}{!}{
\begin{tabular}{l|cccc|cccc}
\toprule

\multicolumn{1}{c|}{\multirow{2}{*}{\bfseries Method}}
& \multicolumn{4}{c|}{
    \bfseries\cellcolor[HTML]{E9F6E9}
    VLN-CE R2R Val-Unseen}
& \multicolumn{4}{c}{
    \bfseries\cellcolor[HTML]{E6F4FF}
    VLN-CE RxR Val-Unseen} \\

\multicolumn{1}{c|}{}
& NE$\downarrow$
& OSR$\uparrow$
& SR$\uparrow$
& SPL$\uparrow$
& NE$\downarrow$
& SR$\uparrow$
& SPL$\uparrow$
& nDTW$\uparrow$ \\

\midrule

\multicolumn{9}{l}{
    \cellcolor[HTML]{FFF4E5}
    \bfseries Supervised Learning (Training-Based Methods)
} \\

NaVid \textcolor{tablegray}{[RSS24]}
& 5.47 & 49.1 & 37.4 & 35.9
& 8.41 & 34.5 & 23.8 & -- \\

Uni-NaVid \textcolor{tablegray}{[RSS25]}
& 5.58 & 53.3 & 47.0 & 42.7
& 6.24 & 48.7 & 40.9 & -- \\

ETPNav \textcolor{tablegray}{[TPAMI24]}
& 4.71 & 65.0 & 57.0 & 49.0
& 5.64 & 54.7 & 44.8 & -- \\

InternVLA-N1 \textcolor{tablegray}{[ICLR26]}
& 4.83 & 63.3 & 58.2 & 54.0
& 5.91 & 53.5 & 46.1 & 65.3 \\

NavFoM \textcolor{tablegray}{[ICLR26]}
& 3.78 & 70.8 & 61.7 & 55.3
& 3.83 & 64.4 & 56.2 & -- \\

ABot-N0 \textcolor{tablegray}{[CVPR26]}
& 3.78 & 70.8 & 66.4 & 63.9
& 3.83 & 69.3 & 60.0 & -- \\

\midrule

\multicolumn{9}{l}{
    \cellcolor[HTML]{FFF4E5}
    \bfseries Zero-Shot (Training-Free Methods)
} \\

Open-Nav$^\dagger$ \textcolor{tablegray}{[ICRA25]}
& 6.70 & 23.0 & 19.0 & 16.1
& -- & -- & -- & -- \\

CA-Nav \textcolor{tablegray}{[TPAMI25]}
& 7.58 & 48.0 & 25.3 & 10.8
& 10.40 & 19.0 & 6.0 & 13.5 \\

InstructNav$^\dagger$ \textcolor{tablegray}{[CoRL24]}
& 6.89 & 47.0 & 31.0 & 24.0
& -- & -- & -- & -- \\

GC-VLN \textcolor{tablegray}{[CoRL25]}
& 7.30 & 41.8 & 33.6 & 16.3
& 8.80 & 33.8 & 13.8 & -- \\

Uni-LaViRA-GPT-5.5$^\dagger$ \textcolor{tablegray}{[arXiv26]}
& 6.11 & 58.0 & 42.0 & 25.7
& -- & -- & -- & -- \\

EvoNav-Gemini-2.5-pro$^\dagger$ \textcolor{tablegray}{[CVPR26]}
& 5.04 & 51.0 & 43.0 & 37.8
& -- & -- & -- & -- \\

SmartWay-GPT-5.5$^\dagger$ \textcolor{tablegray}{[IROS25]}
& 5.16 & 60.0 & 44.0 & 35.0
& -- & -- & -- & -- \\

HSGM-GPT-5 \textcolor{tablegray}{[CVPR26]}
& 5.42 & 58.7 & 47.9 & 32.8
& 7.43 & 41.8 & 25.1 & 54.9 \\

mini-swe-agent-GPT-5.5$^\dagger$ \textcolor{tablegray}{[arXiv26]}
& 7.29 & 57.0 & 52.0 & 44.2
& -- & -- & -- & -- \\

AgenticNav-GPT-5.5$^\dagger$ \textcolor{tablegray}{[arXiv26]}
& 5.19 & 65.0 & 55.0 & 48.4
& -- & -- & -- & -- \\

\rowcolor[HTML]{F4F0FF}
\textbf{HarnessVLN-GPT-5.5$^\dagger$ (Ours)}
& \textbf{4.24}
& \textbf{77.0}
& \textbf{58.0}
& \textbf{35.8}
& -- & -- & -- & --\\

\rowcolor[HTML]{F4F0FF}
\textbf{HarnessVLN-GPT-5.5 (Ours)}
& \textbf{4.08}
& \textbf{70.7}
& \textbf{59.6}
& \textbf{43.6}
& \textbf{6.50}
& \textbf{51.4}
& \textbf{38.1}
& \textbf{55.6}\\

\bottomrule
\end{tabular}
}
\end{table*}

\begin{table}[t!]
\centering
\caption{Comparison with existing methods on HM3D-v2 and HM3D-OVON.
Training indicates whether task-specific navigation training is required.}
\label{tab:object_navigation}

\resizebox{0.7\textwidth}{!}{
\begin{tabular}{l|c|cc|cc}
\toprule
\multicolumn{1}{c|}{\multirow{2}{*}{\bfseries Method}}
& \multicolumn{1}{c|}{\multirow{2}{*}{\bfseries Training}}
& \multicolumn{2}{c|}{
    \bfseries\cellcolor[HTML]{E9F6E9} HM3D-v2}
& \multicolumn{2}{c}{
    \bfseries\cellcolor[HTML]{E6F4FF} HM3D-OVON} \\

\multicolumn{1}{c|}{}
& \multicolumn{1}{c|}{}
& SR$\uparrow$
& SPL$\uparrow$
& SR$\uparrow$
& SPL$\uparrow$ \\

\midrule

Uni-NaVid \textcolor{tablegray}{[RSS25]}
& \cmark & 73.7 & 37.1 & 39.5 & 19.8 \\

FiLM-Nav \textcolor{tablegray}{[arXiv25]}
& \cmark & 77.0 & 41.3 & 40.8 & 24.4 \\

MTU3D \textcolor{tablegray}{[ICCV25]}
& \cmark & -- & -- & 40.8 & 12.1 \\

NavFoM \textcolor{tablegray}{[ICLR26]}
& \cmark & -- & -- & 43.6 & 31.3 \\

ABot-N0 \textcolor{tablegray}{[CVPR26]}
& \cmark & -- & -- & 54.0 & 30.5 \\

\midrule

InstructNav \textcolor{tablegray}{[CoRL24]}
& \xmark & 58.0 & 20.9 & -- & -- \\

TANGO \textcolor{tablegray}{[ICRA25]}
& \xmark & -- & -- & 35.5 & 19.5 \\

VLFM \textcolor{tablegray}{[ICRA24]}
& \xmark & 63.6 & 32.5 & 38.5 & 22.2 \\

STEGNav \textcolor{tablegray}{[arXiv26]}
& \xmark & 69.4 & 28.2 & -- & -- \\

DRIVE-Nav \textcolor{tablegray}{[arXiv26]}
& \xmark & 72.4 & 41.3 & 50.2 & 32.6 \\

MSGNav \textcolor{tablegray}{[CVPR26]}
& \xmark & 74.4 & 33.4 & 48.3 & 27.0 \\

\rowcolor[HTML]{F4F0FF}
\textbf{HarnessVLN (Ours)}
& \xmark
& \textbf{76.0}
& \textbf{37.9}
& \textbf{59.3}
& \textbf{36.6} \\

\bottomrule
\end{tabular}
}
\end{table}

\section{Experiments}
\label{sec:experiments}

We evaluate HarnessVLN on instruction-following and object-goal navigation without task-specific navigation training, using cumulative ablations to assess its design's impact on success, efficiency, and termination behavior. We further validate the framework on a humanoid robot.

\subsection{Experimental Setup}
\label{sec:experimental_setup}

\paragraph{Instruction-following navigation.}
We evaluate on two standard VLN-CE benchmarks, R2R-CE and RxR-CE, using their full val-unseen splits following prior work. For consistency with prior training-free navigation methods~\citep{opennav,agenticnav}, we additionally report results on the R2R-CE Rand100 subset, which we also use for component ablations. We report Success Rate (SR), Success weighted by Path Length (SPL), Oracle Success Rate (OSR), normalized Dynamic Time Warping (nDTW), and Navigation Error (NE).

\paragraph{Object-goal navigation.}
We evaluate on the HM3D-v2 ObjectNav and HM3D-OVON validation splits, containing 1,000 and 3,000 episodes, respectively, in HM3D-Semantics v0.2 scenes. HM3D-v2 uses fixed object categories, whereas HM3D-OVON evaluates open-vocabulary targets.

\subsection{Implementation Details}
\label{sec:implementation_details}

The agent receives synchronized RGB and depth observations at a resolution of $640\!\times\!480$ with a $79^{\circ}$ horizontal field of view. 
GroundingDINO~\citep{groundingdino} and SAM~\citep{sam} provide open-vocabulary target regions and segmentation masks, respectively. The Navigation Executor uses an FMM planner for planar motion and invokes NavDP~\citep{navdp} when stair traversal is required. We use \texttt{GPT-5.5} as the base model for instruction-following navigation and \texttt{GPT-5.6-luna} for object-goal navigation.

\subsection{Main Experiments}
\label{sec:main_results}

As shown in Table~\ref{tab:instruction_navigation}, HarnessVLN achieves SRs of 59.6\% and 51.4\% on the full R2R and RxR val-unseen splits, exceeding HSGM by 11.7 and 9.6 percentage points, respectively. A single \texttt{GPT-5.5} backbone supports all reasoning functions, yielding an OSR of 70.7\% on full R2R. On R2R Rand100, HarnessVLN achieves 58.0\% SR and 77.0\% OSR, outperforming AgenticNav with the same backbone by 3.0 and 12.0 percentage points, respectively, while reducing NE from 5.19\,m to 4.24\,m. These matched-model results support the effectiveness of the navigation-specific harness in improving task completion and goal reaching.

For ObjectNav, HarnessVLN achieves the highest training-free SR on HM3D-v2 at 76.0\%, surpassing MSGNav by 1.6 points and approaching the best listed training-based result of 77.0\% (Table~\ref{tab:object_navigation}). On HM3D-OVON, it achieves the highest SR and SPL among all listed methods at 59.3\% and 36.6\%, outperforming DRIVE-Nav by 9.1 and 4.0 points, respectively. Its SR also exceeds the strongest listed training-based result, achieved by ABot-N0, by 5.3 points. These results demonstrate strong task completion across both navigation families through a shared Harness protocol without task-specific navigation training, although efficiency gains vary across benchmarks.

\subsection{Ablation Studies}
\label{sec:ablation}

We cumulatively enable hierarchical event memory (Mem.), the ST Graph (Graph), and stop validation (Stop) on fixed 100-episode subsets of R2R and HM3D-OVON, keeping episodes, base models, and evaluation settings identical across variants within each task. Subset coverage and episode selection are detailed in the appendix.
As shown in Table~\ref{tab:component_ablation}, event memory improves SR by 7.0 percentage points on both tasks. The ST Graph further increases SR by 6.0 and 1.0 points on R2R and HM3D-OVON, respectively, while improving SPL on both subsets. Stop validation adds another 2.0 SR points on both tasks. Overall, the full Harness improves SR by 15.0 and 10.0 points over the base agent, with gains in both SPL and OSR. These cumulative comparisons measure each component's incremental contribution given those already enabled.

\begin{table*}[t]
    \centering
    \caption{Cumulative ablation of HarnessVLN on fixed 100-episode subsets.}
    \label{tab:component_ablation}
    \resizebox{0.85\textwidth}{!}{%
    \begin{tabular}{ccc ccc ccc}
        \toprule
        \multicolumn{3}{c}{\textbf{Harness Components}} &
        \multicolumn{3}{c}{\textbf{Instruction Following (R2R)}} &
        \multicolumn{3}{c}{\textbf{ObjectNav (HM3D-OVON)}} \\
        \cmidrule(lr){1-3}
        \cmidrule(lr){4-6}
        \cmidrule(lr){7-9}
        Mem. & Graph & Stop &
        SR $\uparrow$ & SPL $\uparrow$ & OSR $\uparrow$ &
        SR $\uparrow$ & SPL $\uparrow$ & OSR $\uparrow$ \\
        \midrule
        \xmark & \xmark & \xmark &
        43.0 & 24.3 & 65.0 & 45.0 & 32.0 & 72.0 \\
        \cmark & \xmark & \xmark &
        50.0 & 30.9 & 73.0 & 52.0 & 32.4 & 72.0 \\
        \cmark & \cmark & \xmark &
        56.0 & 32.4 & 76.0 & 53.0 & 34.2 & 71.0 \\
        \midrule
        \rowcolor[HTML]{F4F0FF}
        \cmark & \cmark & \cmark &
        \textbf{58.0} & \textbf{35.9} & \textbf{77.0} &
        \textbf{55.0} & \textbf{33.0} & \textbf{74.0} \\
        \bottomrule
    \end{tabular}}
\end{table*}

\section{Real-World Deployment}
\label{sec:real_world}

We deploy HarnessVLN on a humanoid robot. All tasks use the same Harness protocol, without training task-specific policies in the deployment environment.

\subsection{Robot Platform and System Configuration}
\label{sec:robot_platform}
AGIBOT A3 Ultra is a full-size humanoid robot measuring 1.74\,m in height. It is equipped with a 3D LiDAR and multiple RGB-D and fisheye cameras, with an onboard computing architecture based on NVIDIA Thor. We use the head-mounted stereo cameras as the primary visual sensors. Fast-FoundationStereo~\citep{fastfoundationstereo} estimates dense metric depth from synchronized stereo images. Qwen-3.8-27B serves as the planning model within the Harness, while a local path planner converts validated spatial goals into executable waypoints and motion commands.

\begin{figure}[t]
  \centering
  \includegraphics[width=\linewidth]{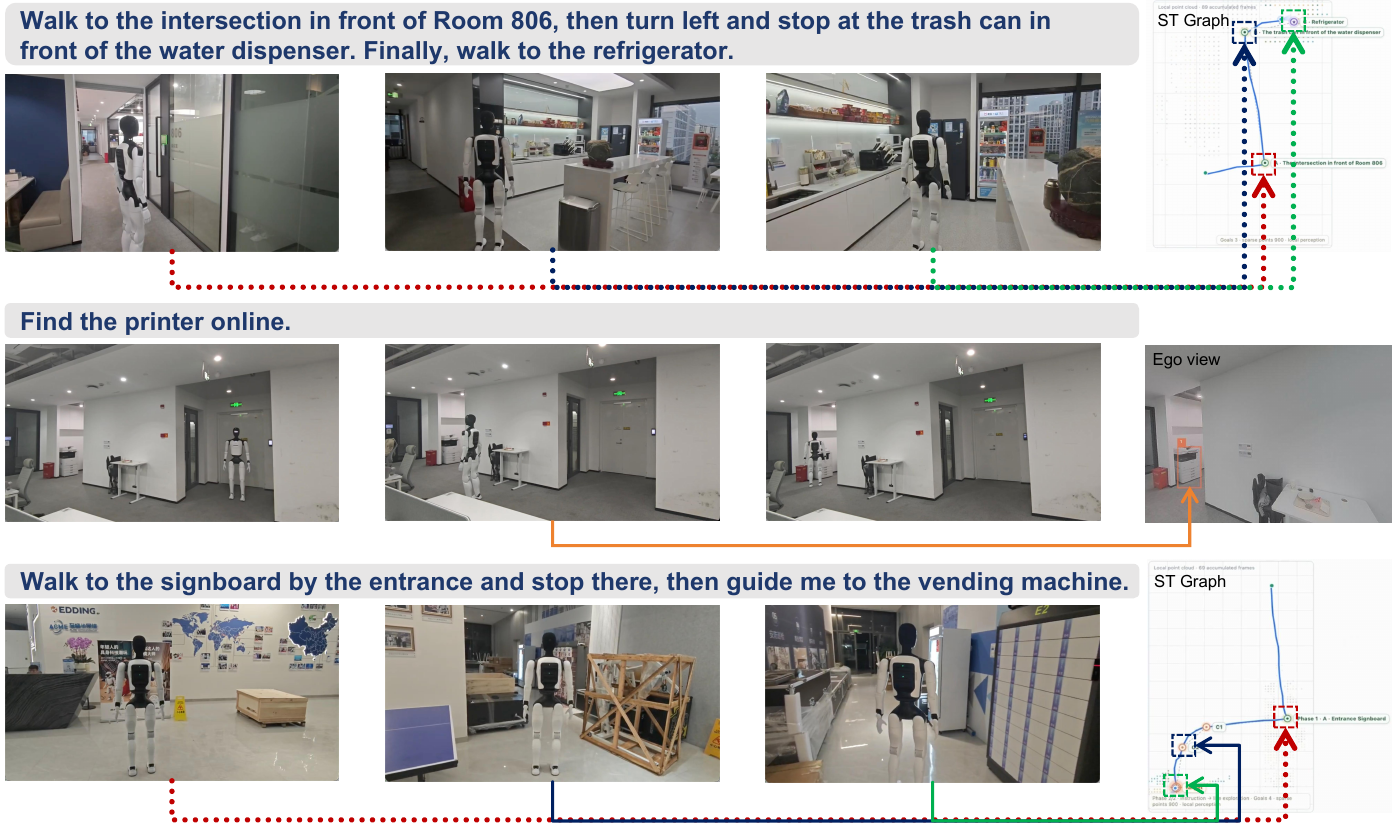}
  \caption{Real-world navigation examples: sequential instruction following, open-vocabulary object search, and a combined task involving route following and object search. }
  \label{fig:realworld_vis}
\end{figure}

\subsection{Real-World Navigation Experiments}
\label{sec:real_world_eval}

We evaluate HarnessVLN in three real-world settings (Figure~\ref{fig:realworld_vis}): instruction following through the intersection outside Room 806, followed by a left turn and a stop at the trash can near the water dispenser, before proceeding to the refrigerator; open-vocabulary search for a printer; and a combined task that first requires stopping at the entrance signboard and then locating a vending machine. Across these settings, the shared Harness tracks subgoal progress, grounds landmarks and targets in visual observations, and validates stopping requests using semantic and spatial evidence. The ST Graph retains landmark and target evidence across navigation stages, supporting continuity between route following and object search. HarnessVLN manages task progress, spatial evidence, and stop validation through a shared runtime protocol.

\section{Conclusion}
\label{sec:conclusion}

We presented HarnessVLN, a training-free framework that unifies instruction-following and object-goal navigation through a shared Agent Harness. By coordinating MLLM planning, hierarchical memory, spatial evidence, and validated tool execution, HarnessVLN bridges semantic reasoning and grounded action. Benchmark results demonstrate improved task completion across both tasks, while humanoid deployment shows real-world applicability. The current Harness relies on predefined orchestration and validation; future work will explore self-evolving mechanisms refined through interaction in open environments.

\begingroup
\setlength{\bibsep}{2pt plus 1pt}
\bibliography{references}

@inproceedings{cow,
  author = {Gadre, Samir Yitzhak and Wortsman, Mitchell and Ilharco, Gabriel and Schmidt, Ludwig and Song, Shuran},
  title = {{CoWs} on Pasture: Baselines and Benchmarks for Language-Driven Zero-Shot Object Navigation},
  booktitle = {Proceedings of the IEEE/CVF Conference on Computer Vision and Pattern Recognition},
  pages = {23171--23181},
  year = {2023}
}

@inproceedings{esc,
  author = {Zhou, Kaiwen and Zheng, Kaizhi and Pryor, Connor and Shen, Yilin and Jin, Hongxia and Getoor, Lise and Wang, Xin Eric},
  title = {{ESC}: Exploration with Soft Commonsense Constraints for Zero-Shot Object Navigation},
  booktitle = {Proceedings of the 40th International Conference on Machine Learning},
  pages = {42829--42842},
  year = {2023}
}

@inproceedings{vlfm,
  author = {Yokoyama, Naoki and Ha, Sehoon and Batra, Dhruv and Wang, Jiuguang and Bucher, Bernadette},
  title = {{VLFM}: Vision-Language Frontier Maps for Zero-Shot Semantic Navigation},
  booktitle = {Proceedings of the IEEE International Conference on Robotics and Automation},
  pages = {42--48},
  year = {2024}
}

@article{apexnav,
  author = {Zhang, Mingjie and Du, Yuheng and Wu, Chengkai and Zhou, Jinni and Qi, Zhenchao and Ma, Jun and Zhou, Boyu},
  title = {{ApexNav}: An Adaptive Exploration Strategy for Zero-Shot Object Navigation with Target-centric Semantic Fusion},
  journal = {IEEE Robotics and Automation Letters},
  volume = {10},
  number = {11},
  pages = {11530--11537},
  year = {2025}
}

@inproceedings{instructnav,
  author = {Long, Yuxing and Cai, Wenzhe and Wang, Hongcheng and Zhan, Guanqi and Dong, Hao},
  title = {{InstructNav}: Zero-Shot System for Generic Instruction Navigation in Unexplored Environment},
  booktitle = {Proceedings of the 8th Conference on Robot Learning},
  pages = {2049--2060},
  year = {2025}
}

@inproceedings{opennav,
  author = {Qiao, Yanyuan and Lyu, Wenqi and Wang, Hui and Wang, Zixu and Li, Zerui and Zhang, Yuan and Tan, Mingkui and Wu, Qi},
  title = {{Open-Nav}: Exploring Zero-Shot Vision-and-Language Navigation in Continuous Environment with Open-Source {LLMs}},
  booktitle = {Proceedings of the IEEE International Conference on Robotics and Automation},
  pages = {6710--6717},
  year = {2025}
}

@article{abot,
  author = {Chu, Zedong and Xie, Shichao and Wu, Xiaolong and Shen, Yanfen and Luo, Minghua and Wang, Zhengbo and Liu, Fei and Leng, Xiaoxu and Hu, Junjun and Yin, Mingyang and others},
  title = {{ABot-N0}: Technical Report on the {VLA} Foundation Model for Versatile Embodied Navigation},
  journal = {arXiv preprint arXiv:2602.11598},
  year = {2026}
}

@inproceedings{omninav,
  author = {Xue, Xinda and Hu, Junjun and Luo, Minghua and Xie, Shichao and Chen, Jintao and Xie, Zixun and Quan, Kuichen and Guo, Wei and Chu, Zedong and Xu, Mu and others},
  title = {{OmniNav}: A Unified Framework for Prospective Exploration and Visual-Language Navigation},
  booktitle = {Proceedings of the 14th International Conference on Learning Representations},
  year = {2026}
}

@inproceedings{streamvln,
  author = {Wei, Meng and Wan, Chenyang and Yu, Xiqian and Wang, Tai and Yang, Yuqiang and Mao, Xiaohan and Zhu, Chenming and Cai, Wenzhe and Wang, Hanqing and Chen, Yilun and others},
  title = {{StreamVLN}: Streaming Vision-and-Language Navigation via {SlowFast} Context Modeling},
  booktitle = {Proceedings of the IEEE International Conference on Robotics and Automation},
  year = {2026}
}

@inproceedings{navila,
  author = {Cheng, An-Chieh and Ji, Yandong and Yang, Zhaojing and Gongye, Zaitian and Zou, Xueyan and Kautz, Jan and Biyik, Erdem and Yin, Hongxu and Liu, Sifei and Wang, Xiaolong},
  title = {{NaVILA}: Legged Robot Vision-Language-Action Model for Navigation},
  booktitle = {Proceedings of Robotics: Science and Systems},
  year = {2025}
}

@inproceedings{janusvln,
  author = {Zeng, Shuang and Qi, Dekang and Chang, Xinyuan and Xiong, Feng and Xie, Shichao and Wu, Xiaolong and Liang, Shiyi and Xu, Mu and Wei, Xing},
  title = {{JanusVLN}: Decoupling Semantics and Spatiality with Dual Implicit Memory for Vision-Language Navigation},
  booktitle = {Proceedings of the 14th International Conference on Learning Representations},
  year = {2026}
}

@inproceedings{navid,
  author = {Zhang, Jiazhao and Wang, Kunyu and Xu, Rongtao and Zhou, Gengze and Hong, Yicong and Fang, Xiaomeng and Wu, Qi and Zhang, Zhizheng and Wang, He},
  title = {{NaVid}: Video-Based {VLM} Plans the Next Step for Vision-and-Language Navigation},
  booktitle = {Proceedings of Robotics: Science and Systems},
  year = {2024}
}

@inproceedings{uninavid,
  author = {Zhang, Jiazhao and Wang, Kunyu and Wang, Shaoan and Li, Minghan and Liu, Haoran and Wei, Songlin and Wang, Zhongyuan and Zhang, Zhizheng and Wang, He},
  title = {{Uni-NaVid}: A Video-Based Vision-Language-Action Model for Unifying Embodied Navigation Tasks},
  booktitle = {Proceedings of Robotics: Science and Systems},
  year = {2025}
}

@inproceedings{navfom,
  author = {Zhang, Jiazhao and Li, Anqi and Qi, Yunpeng and Li, Minghan and Liu, Jiahang and Wang, Shaoan and Liu, Haoran and Zhou, Gengze and Wu, Yuze and Li, Xingxing and Fan, Yuxin and Li, Wenjun and Chen, Zhibo and Gao, Fei and Wu, Qi and Zhang, Zhizheng and Wang, He},
  title = {Embodied Navigation Foundation Model},
  booktitle = {Proceedings of the 14th International Conference on Learning Representations},
  year = {2026}
}

@article{aspire,
  author = {Lu, Runyu and Wu, Yubo and Kou, Ethan and Fu, Letian and Xiao, Wenli and Mandlekar, Ajay and Xu, Yinzhen and Shi, Guanya and Goldberg, Ken and Chen, Ang and others},
  title = {{ASPIRE}: Agentic /Skills Discovery for Robotics},
  journal = {arXiv preprint arXiv:2607.00272},
  year = {2026}
}

@article{harnessvla,
  author = {Zhang, Yixian and Zhang, Huanming and Gao, Feng and Li, Xiao and Liu, Zhihao and Zhu, Chunyang and Qiu, Jiaxing and Yan, Yuchen and Liu, Jiyuan and Tang, Wenhao and others},
  title = {{Harness VLA}: Steering Frozen {VLAs} into Reliable Manipulation Primitives via Memory-Guided Agents},
  journal = {arXiv preprint arXiv:2607.08448},
  year = {2026}
}

@inproceedings{codeaspolicies,
  author = {Liang, Jacky and Huang, Wenlong and Xia, Fei and Xu, Peng and Hausman, Karol and Ichter, Brian and Florence, Pete and Zeng, Andy},
  title = {{Code as Policies}: Language Model Programs for Embodied Control},
  booktitle = {Proceedings of the IEEE International Conference on Robotics and Automation},
  pages = {9493--9500},
  year = {2023}
}

@inproceedings{voxposer,
  author = {Huang, Wenlong and Wang, Chen and Zhang, Ruohan and Li, Yunzhu and Wu, Jiajun and Fei-Fei, Li},
  title = {{VoxPoser}: Composable {3D} Value Maps for Robotic Manipulation with Language Models},
  booktitle = {Proceedings of the 7th Conference on Robot Learning},
  pages = {540--562},
  year = {2023}
}

@inproceedings{cap-x,
  author = {Fu, Max and Yu, Justin and El-Refai, Karim and Kou, Ethan and Xue, Haoru and Huang, Huang and Xiao, Wenli and Wang, Guanzhi and Li, Fei-Fei and others},
  title = {{CaP-X}: A Framework for Benchmarking and Improving Coding Agents for Robot Manipulation},
  booktitle = {Proceedings of the 43rd International Conference on Machine Learning},
  year = {2026}
}

@inproceedings{saycan,
  author = {Ichter, Brian and Brohan, Anthony and Chebotar, Yevgen and Finn, Chelsea and Hausman, Karol and Herzog, Alexander and Ho, Daniel and Ibarz, Julian and Irpan, Alex and Jang, Eric and Julian, Ryan and Kalashnikov, Dmitry and Levine, Sergey and Lu, Yao and Parada, Carolina and Rao, Kanishka and Sermanet, Pierre and Toshev, Alexander T. and Vanhoucke, Vincent and Xia, Fei and Xiao, Ted and Xu, Peng and Yan, Mengyuan and Brown, Noah and Ahn, Michael and Cortes, Omar and Sievers, Nicolas and Tan, Clayton and Xu, Sichun and Reyes, Diego and Rettinghouse, Jarek and Quiambao, Jornell and Pastor, Peter and Luu, Linda and Lee, Kuang-Huei and Kuang, Yuheng and Jesmonth, Sally and Joshi, Nikhil J. and Jeffrey, Kyle and Ruano, Rosario Jauregui and Hsu, Jasmine and Gopalakrishnan, Keerthana and David, Byron and Zeng, Andy and Fu, Chuyuan Kelly},
  title = {Do As I Can, Not As I Say: Grounding Language in Robotic Affordances},
  booktitle = {Proceedings of the 6th Conference on Robot Learning},
  pages = {287--318},
  year = {2023}
}

@inproceedings{roboclaw,
  author = {Li, Ruiying and Zhou, Yunlang and Zhu, YuYao and Chen, Kylin and Wang, Jingyuan and Wang, Sukai and Hu, Kongtao and Yu, Minhui and Jiang, Bowen and Su, Zhan and others},
  title = {{RoboClaw}: An Agentic Framework for Scalable Long-Horizon Robotic Tasks},
  booktitle = {Proceedings of the European Conference on Computer Vision},
  year = {2026}
}

@article{agenticrobot,
  author = {Yang, Zhejian and Chen, Yongchao and Zhou, Xueyang and Yan, Jiangyue and Song, Dingjie and Liu, Yinuo and Li, Yuting and Zhang, Yu and Zhou, Pan and Chen, Hechang and others},
  title = {{Agentic Robot}: A Brain-Inspired Framework for Vision-Language-Action Models in Embodied Agents},
  journal = {arXiv preprint arXiv:2505.23450},
  year = {2025}
}

@inproceedings{mtu3d,
  author = {Zhu, Ziyu and Wang, Xilin and Li, Yixuan and Zhang, Zhuofan and Ma, Xiaojian and Chen, Yixin and Jia, Baoxiong and Liang, Wei and Yu, Qian and Deng, Zhidong and others},
  title = {Move to Understand a {3D} Scene: Bridging Visual Grounding and Exploration for Efficient and Versatile Embodied Navigation},
  booktitle = {Proceedings of the IEEE/CVF International Conference on Computer Vision},
  pages = {8120--8132},
  year = {2025}
}

@article{etpnav,
  author = {An, Dong and Wang, Hanqing and Wang, Wenguan and Wang, Zun and Huang, Yan and He, Keji and Wang, Liang},
  title = {{ETPNav}: Evolving Topological Planning for Vision-Language Navigation in Continuous Environments},
  journal = {IEEE Transactions on Pattern Analysis and Machine Intelligence},
  volume = {47},
  number = {7},
  pages = {5130--5145},
  year = {2025}
}

@inproceedings{mapdream,
  author = {Lian, Guoxin and Wang, Shuo and Wang, Yucheng and Wang, Yongcai and Chen, Maiyue and Wang, Kaihui and Zhang, Bo and Su, Zhizhong and Li, Deying and Fan, Zhaoxin},
  title = {{MapDream}: Task-Driven Map Learning for Vision-Language Navigation},
  booktitle = {Proceedings of the 43rd International Conference on Machine Learning},
  year = {2026}
}

@article{canav,
  author = {Chen, Kehan and An, Dong and Huang, Yan and Xu, Rongtao and Su, Yifei and Ling, Yonggen and Reid, Ian and Wang, Liang},
  title = {Constraint-Aware Zero-Shot Vision-Language Navigation in Continuous Environments},
  journal = {IEEE Transactions on Pattern Analysis and Machine Intelligence},
  volume = {47},
  number = {11},
  pages = {10441--10456},
  year = {2025}
}

@inproceedings{hsgm,
  author = {Li, Kailing and Qian, Tianwen and Yang, Lijin and Fu, Yuqian and Gong, Jingyu and Wang, Xiaoling and He, Liang},
  title = {Bridging the {2D-3D} Gap: A Hierarchical Semantic-Geometric Map for Vision Language Navigation},
  booktitle = {Proceedings of the IEEE/CVF Conference on Computer Vision and Pattern Recognition},
  pages = {15243--15252},
  year = {2026}
}

@article{filmnav,
  author = {Yokoyama, Naoki and Ha, Sehoon},
  title = {{FiLM-Nav}: Efficient and Generalizable Navigation via {VLM} Fine-Tuning},
  journal = {arXiv preprint arXiv:2509.16445},
  year = {2025}
}

@inproceedings{tango,
  author = {Podgorski, Stefan and Garg, Sourav and Hosseinzadeh, Mehdi and Mares, Lachlan and Dayoub, Feras and Reid, Ian},
  title = {{Tango}: Traversability-Aware Navigation with Local Metric Control for Topological Goals},
  booktitle = {Proceedings of the IEEE International Conference on Robotics and Automation},
  pages = {2399--2406},
  year = {2025}
}

@article{drivenav,
  author = {Gao, Maoguo and Zhu, Zejun and Sun, Zhiming and Ma, Zhengwei and Yuan, Longze and Ma, Zhongjing and Gao, Zhigang and Zhang, Jinhui and Zou, Suli},
  title = {{DRIVE-Nav}: Directional Reasoning, Inspection, and Verification for Efficient Open-Vocabulary Navigation},
  journal = {IEEE Robotics and Automation Letters},
  volume = {11},
  pages = {11158--11165},
  year = {2026}
}

@inproceedings{msgnav,
  author = {Huang, Xun and Zhao, Shijia and Wang, Yunxiang and Lu, Xin and Zhang, Wanfa and Qu, Rongsheng and Li, Weixin and Wang, Yunhong and Wen, Chenglu},
  title = {{MSGNav}: Unleashing the Power of Multi-modal {3D} Scene Graph for Zero-Shot Embodied Navigation},
  booktitle = {Proceedings of the IEEE/CVF Conference on Computer Vision and Pattern Recognition},
  pages = {37154--37163},
  year = {2026}
}

@inproceedings{ground,
  author = {Wei, Meng and Wan, Chenyang and Peng, Jiaqi and Yu, Xiqian and Yang, Yuqiang and Feng, Delin and Cai, Wenzhe and Zhu, Chenming and Wang, Tai and Pang, Jiangmiao and Liu, Xihui},
  title = {{Ground Slow, Move Fast}: A Dual-System Foundation Model for Generalizable Vision-and-Language Navigation},
  booktitle = {Proceedings of the 14th International Conference on Learning Representations},
  year = {2026}
}

@inproceedings{lavira,
  author = {Ding, Hongyu and Xu, Ziming and Fang, Yudong and Wu, You and Chen, Zixuan and Shi, Jieqi and Huo, Jing and Zhang, Yifan and Gao, Yang},
  title = {{LaViRA}: Language-Vision-Robot Actions Translation for Zero-Shot Vision Language Navigation in Continuous Environments},
  booktitle = {Proceedings of the IEEE International Conference on Robotics and Automation},
  year = {2026}
}

@article{unilavira,
  author = {Ding, Hongyu and Zhang, Sizhuo and Xu, Ziming and Guo, Jinwen and Liu, Hongxiu and Cheng, Xingzhi and Chen, Zixuan and Qi, Haifei and Wang, Duo and Xu, Hao and others},
  title = {{Uni-LaViRA}: Language-Vision-Robot Actions Translation for Unified Embodied Navigation},
  journal = {arXiv preprint arXiv:2605.27582},
  year = {2026}
}

@article{agenticnav,
  author = {Li, Yijian and Li, Changze and Shi, Hantian and Luo, Jiaying and Cai, Jiyuan and Yang, Ming and Qin, Tong},
  title = {{AgenticNav}: Zero-Shot Vision-and-Language Navigation as a Tool-Calling Harness},
  journal = {arXiv preprint arXiv:2606.10577},
  year = {2026}
}

@inproceedings{hm3dsem,
  author = {Yadav, Karmesh and Ramrakhya, Ram and Ramakrishnan, Santhosh Kumar and Gervet, Theo and Turner, John and Gokaslan, Aaron and Maestre, Noah and Chang, Angel Xuan and Batra, Dhruv and Savva, Manolis and others},
  title = {{Habitat-Matterport} {3D} Semantics Dataset},
  booktitle = {Proceedings of the IEEE/CVF Conference on Computer Vision and Pattern Recognition},
  pages = {4927--4936},
  year = {2023}
}

@inproceedings{hm3dovon,
  author = {Yokoyama, Naoki and Ramrakhya, Ram and Das, Abhishek and Batra, Dhruv and Ha, Sehoon},
  title = {{HM3D-OVON}: A Dataset and Benchmark for Open-Vocabulary Object Goal Navigation},
  booktitle = {Proceedings of the IEEE/RSJ International Conference on Intelligent Robots and Systems},
  pages = {5543--5550},
  year = {2024}
}

@inproceedings{sam,
  author = {Kirillov, Alexander and Mintun, Eric and Ravi, Nikhila and Mao, Hanzi and Rolland, Chloe and Gustafson, Laura and Xiao, Tete and Whitehead, Spencer and Berg, Alexander C. and Lo, Wan{-}Yen and others},
  title = {Segment Anything},
  booktitle = {Proceedings of the IEEE/CVF International Conference on Computer Vision},
  pages = {4015--4026},
  year = {2023}
}

@inproceedings{groundingdino,
  author = {Liu, Shilong and Zeng, Zhaoyang and Ren, Tianhe and Li, Feng and Zhang, Hao and Yang, Jie and Jiang, Qing and Li, Chunyuan and Yang, Jianwei and Su, Hang and others},
  title = {{Grounding DINO}: Marrying {DINO} with Grounded Pre-Training for Open-Set Object Detection},
  booktitle = {Proceedings of the European Conference on Computer Vision},
  pages = {38--55},
  year = {2024}
}

@inproceedings{r2r,
  author = {Anderson, Peter and Wu, Qi and Teney, Damien and Bruce, Jake and Johnson, Mark and S{\"u}nderhauf, Niko and Reid, Ian and Gould, Stephen and van den Hengel, Anton},
  title = {Vision-and-Language Navigation: Interpreting Visually-Grounded Navigation Instructions in Real Environments},
  booktitle = {Proceedings of the IEEE Conference on Computer Vision and Pattern Recognition},
  pages = {3674--3683},
  year = {2018}
}

@inproceedings{rxr,
  author = {Ku, Alexander and Anderson, Peter and Patel, Roma and Ie, Eugene and Baldridge, Jason},
  title = {{Room-Across-Room}: Multilingual Vision-and-Language Navigation with Dense Spatiotemporal Grounding},
  booktitle = {Proceedings of the 2020 Conference on Empirical Methods in Natural Language Processing},
  pages = {4392--4412},
  year = {2020}
}

@inproceedings{navdp,
  author = {Cai, Wenzhe and Peng, Jiaqi and Yang, Yuqiang and Zhang, Yujian and Wei, Meng and Wang, Hanqing and Chen, Yilun and Wang, Tai and Pang, Jiangmiao},
  title = {{NavDP}: Learning Sim-to-Real Navigation Diffusion Policy with Privileged Information Guidance},
  booktitle = {Proceedings of the IEEE International Conference on Robotics and Automation},
  year = {2026}
}

@inproceedings{fastfoundationstereo,
  author = {Wen, Bowen and Dewan, Shaurya and Birchfield, Stan},
  title = {{Fast-FoundationStereo}: Real-Time Zero-Shot Stereo Matching},
  booktitle = {Proceedings of the IEEE/CVF Conference on Computer Vision and Pattern Recognition},
  pages = {7513--7524},
  year = {2026}
}

@article{stegnav,
  author = {Chen, Yang and Huang, Zhenyu and Fu, Wenbo and Peng, Danyang and Tian, Shi-Yu and Yu, Kun-Yang and Guo, Lan-Zhe},
  title = {{STEGNav}: Spatio-Temporal Event Graph Reasoning for Multimodal Lifelong Object Navigation},
  journal = {arXiv preprint arXiv:2608.28279},
  year = {2026}
}

@inproceedings{lookstep,
  author = {Yu, Kun-Yang and Li, Yingzhe and Xu, Hongyu and Tian, Shi-Yu and Zhou, Zhi and Chen, Yang and Yang, Ming and Wang, Sheng and Yu, Qing and Guo, Lan-Zhe and Li, Yu-Feng},
  title = {{LookStep}: Efficient Vision-Language Navigation with Linguistic Foresight and Event Driven Memory},
  booktitle = {Proceedings of the 2026 Conference on Empirical Methods in Natural Language Processing},
  year = {2026}
}

@inproceedings{re2agent,
  author = {Chen, Yang and You, Hong-Jie and Shao, Jie-Jing and Yang, Xiao-Wen and Yang, Ming and Li, Yu-Feng and Guo, Lan-Zhe},
  title = {{Re$^2$ Agent}: Reflection and Re-execution Agent for Embodied Decision Making},
  booktitle = {NeurIPS 2025 Challenge on Foundation Models for Embodied Agents},
  year = {2025}
}

@inproceedings{lifting,
  title={Lifting Traces to Logic: Programmatic Skill Induction with Neuro-Symbolic Learning for Long-Horizon Agentic Tasks},
  author={Jie-Jing Shao and Haiyan Yin and Yueming Lyu and Xingrui Yu and Lan-Zhe Guo and Ivor W. Tsang and James T. Kwok and Yu-Feng Li},
  booktitle={Proceedings of the 43rd International Conference on Machine Learning},
  year={2026}
}

@article{embodiedagent,
  author = {Zhou, Jian and Zhao, Xunyi and Zhou, Gengze and Li, Zerui and Lin, Sihao and Liu, Jiajun and Wu, Qi},
  title = {Embodied Agents Take Control: Minimal-Interface Zero-Shot Agents Rival Industrial-Scale Policies in Vision-and-Language Navigation},
  journal = {arXiv preprint arXiv:2607.26148},
  year = {2026}
}

@inproceedings{smartway,
  author = {Shi, Xiangyu and Li, Zerui and Lyu, Wenqi and Xia, Jiatong and Dayoub, Feras and Qiao, Yanyuan and Wu, Qi},
  title = {{SmartWay}: Enhanced Waypoint Prediction and Backtracking for Zero-Shot Vision-and-Language Navigation},
  booktitle = {Proceedings of the IEEE/RSJ International Conference on Intelligent Robots and Systems},
  pages = {16923--16930},
  year = {2025}
}

@inproceedings{evonav,
  author = {Dai, Guangzhao and Wang, Shuo and Wang, Zihan and Xie, Guo{-}Sen and Yang, Yang and Pan, Jinshan and Sun, Qianru and Shu, Xiangbo},
  title = {History to Future: Evolving Agent with Experience and Thought for Zero-Shot Vision-and-Language Navigation},
  booktitle = {Proceedings of the IEEE/CVF Conference on Computer Vision and Pattern Recognition},
  pages = {15177--15187},
  year = {2026}
}
\bibliographystyle{plainnat}
\endgroup

\clearpage
\appendix

\setcounter{section}{0}
\renewcommand{\thesection}{\arabic{section}}
\renewcommand{\thesubsection}{\thesection.\arabic{subsection}}

\ifdefined\theHsection
  \renewcommand{\theHsection}{appendix.\arabic{section}}
\fi

\begin{center}
    {\LARGE\bfseries Appendix\par}
\end{center}
\vspace{1em}

\section{Experimental Setup and Implementation Details}
\label{app:settings}

\subsection{Benchmarks and Evaluation Protocol}
\label{app:evaluation_protocol}
Main results use the full evaluation splits of VLN-CE R2R and RxR in Matterport3D, and HM3D-v2 and HM3D-OVON in HM3D-Semantics v0.2. To align with prior evaluation protocols~\citep{opennav,agenticnav}, we additionally report main results on the R2R-CE val-unseen Rand100 subset. RxR uses US English guide instructions. Component ablations are conducted on the R2R val-unseen Rand100 subset and a fixed 100-episode subset of HM3D-OVON val. Rand100 covers all 11 scenes in the full 1,839-episode R2R split, while the HM3D-OVON subset spans 32 of 36 scenes and all 49 target categories in the full 3,000-episode validation split.

\subsection{Models, Tools, and Generation Settings}
\label{app:models_tools}
A single pretrained MLLM handles task decomposition, planning, grounding, stop verification, and recovery through role-specific prompts. The main instruction-following and ObjectNav experiments use \texttt{GPT-5.5} and \texttt{GPT-5.6-luna}, respectively. We adopt selected settings from Uni-LaViRA~\citep{unilavira} and encapsulate GroundingDINO, SAM, FMM, and NavDP as perception and navigation tools within the Agent Harness. No task-specific fine-tuning is performed.

We request a sampling temperature of 0.7 for initialization, planning, verification, and recovery, with output-token limits of 4,096, 8,192, 8,192, and 2,048, respectively. Image-region grounding uses a requested temperature of 0 and an output-token limit of 10,240.

\subsection{Simulation and Harness Configuration}
\label{app:simulation}

\paragraph{Observations and visual context.}
Table~\ref{tab:app_settings} summarizes the configuration. Four-directional observations capture forward, left, behind, and right views at relative headings of $0^\circ$, $90^\circ$, $180^\circ$, and $270^\circ$. Forward-view history is sampled every two environment steps and subsampled to fit the 50-image budget, including current observations. We fix the simulator seed to zero, although stochastic model responses may still cause variation across runs.

\begin{table}[htbp]
\centering
\caption{Simulation and Harness configuration. Shared values apply to both task families.}
\label{tab:app_settings}
\begin{tabularx}{\linewidth}{@{}lXX@{}}
\toprule
Setting & Instruction following & ObjectNav \\
\midrule
Maximum environment steps & 500 & 500 \\
Task success distance & 3.0\,m & 1.0\,m \\
Forward step / turn angle & \multicolumn{2}{l}{0.25\,m / $30^\circ$} \\
Sensor height / agent radius & \multicolumn{2}{l}{0.88\,m / 0.10\,m} \\
Depth interval / map resolution & \multicolumn{2}{l}{$[0.1,5.0]$\,m / 5\,cm} \\
History interval / image budget & \multicolumn{2}{l}{2 steps / 50 images per request} \\
Place / entity merge radius & \multicolumn{2}{l}{0.75\,m / 1.00\,m} \\
Retrieval seeds / place limit & \multicolumn{2}{l}{3 / 5} \\
Recency / salience weights & \multicolumn{2}{l}{0.3 / 0.3} \\
Failure penalty / penalty cap & \multicolumn{2}{l}{0.5 per applicable failure / 1.0} \\
\bottomrule
\end{tabularx}
\end{table}

\paragraph{Memory retrieval.}
The ST Graph computes semantic relevance using local IDF-weighted cosine similarity. Retrieval expands the top three seed places by one graph hop and returns at most five places. Failure penalties depend on the active subgoal and are capped to avoid permanently excluding previously unsuccessful branches.

\paragraph{Geometric stop validation.}
\label{app:stop_validation}
The Harness validates proximity using fresh depth at a grounded region (2.5\,m threshold) or distance to a projected navigation target (1.0\,m). Fresh depth is the median of a $5\times5$ patch at the bounding-box anchor, accepted only when its standard deviation is at most 0.5\,m. Unavailable or unreliable depth triggers the projected-target fallback. These checks do not replace the benchmark's geodesic success criteria.

\paragraph{Step budget and observation overhead.}
We use a default episode limit of 500 environment steps and compare it with an extended 1,000-step limit to assess the effect of the step budget. This analysis uses the same fixed 100-episode R2R-CE val-unseen subset as the Rand100 training-free comparison in Table~\ref{tab:instruction_navigation} and the component ablation in Table~\ref{tab:component_ablation}. A full panorama requires twelve $30^\circ$ rotations, each consuming one environment step. Observation rotations, navigation turns, forward movements, and \texttt{STOP} all count toward the episode limit.

We compare the two budgets by evaluating recorded 1,000-step trajectories and truncating them at 500 steps for offline re-scoring under the same endpoint-distance criterion. The default 500-step budget yields NE $=4.2416$\,m, OSR $=77.0\%$, SR $=58.0\%$, SPL $=35.85\%$, and nDTW $=46.73\%$. Extending the budget to 1,000 steps yields NE $=4.1056$\,m, OSR $=81.0\%$, SR $=61.0\%$, SPL $=35.63\%$, and nDTW $=45.30\%$. The larger budget therefore increases OSR and SR by 4.0 and 3.0 percentage points, respectively, while slightly reducing SPL and nDTW. 

\begin{figure}[htbp]
\centering
\includegraphics[width=\linewidth]{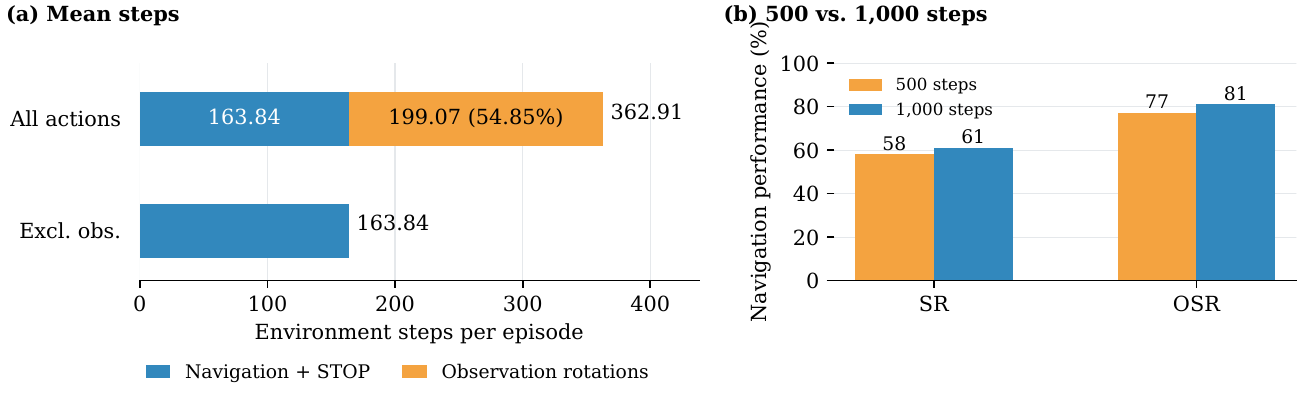}
\caption{(a) Observation rotations account for 54.85\% of all environment steps on R2R-CE Rand100. (b) Navigation performance under 500- and 1,000-step caps on the same subset; the 500-step values are obtained by offline trajectory re-scoring.}
\label{fig:app_step_budget}
\end{figure}

Figure~\ref{fig:app_step_budget} further breaks down step usage in the 1,000-step runs. Observation rotations account for 19,907 of the 36,291 executed actions (54.85\%), averaging 199.07 steps per episode. Excluding these rotations from the action count reduces the mean from 362.91 to 163.84 steps, with navigation turns, forward movements, and \texttt{STOP} still included. The rotation share is calculated over pooled actions across all episodes.

\subsection{Model Sensitivity and Matched-Model Comparison}
\label{sec:model_api_ablation}

We evaluate \texttt{GPT-5.6-luna}, \texttt{GPT-6-astra}, and \texttt{Qwen3.8-flash} on fixed 100-episode subsets of HM3D-v2 and HM3D-OVON, holding the Harness protocol and tools constant. We also re-evaluate MSGNav with \texttt{GPT-5.6-luna} on the same episodes within each benchmark for matched-model comparisons.

Table~\ref{tab:model_api_ablation} shows SRs of 72.0--86.0\% on HM3D-v2 and 51.0--63.0\% on HM3D-OVON, with the Astra variant achieving the highest SR on both subsets. With the same \texttt{GPT-5.6-luna} backbone, HarnessVLN outperforms MSGNav by 14.0 percentage points in SR and 13.1 points in SPL on HM3D-v2, and by 18.0 and 14.1 points, respectively, on HM3D-OVON. The OSR--SR gap narrows from 22.0 to 10.0 points on HM3D-v2, while the larger gap on HM3D-OVON indicates remaining room to improve successful termination.

\begin{table}[htbp]
\centering
\caption{Model sensitivity and matched-model comparison on fixed 100-episode ObjectNav subsets. SR, SPL, and OSR are percentages; Gap is OSR minus SR in percentage points.}
\label{tab:model_api_ablation}
\small
\setlength{\tabcolsep}{3pt}
\begin{tabular}{@{}ll*{8}{r}@{}}
\toprule
\multirow{2}{*}{Method} & \multirow{2}{*}{Model} &
\multicolumn{4}{c}{\textbf{HM3D-v2}} &
\multicolumn{4}{c}{\textbf{HM3D-OVON}} \\
\cmidrule(lr){3-6}
\cmidrule(lr){7-10}
& & SR $\uparrow$ & SPL $\uparrow$ & OSR $\uparrow$ & Gap $\downarrow$ &
SR $\uparrow$ & SPL $\uparrow$ & OSR $\uparrow$ & Gap $\downarrow$ \\
\midrule
MSGNav & \texttt{GPT-5.6-luna} &
62.0 & 27.7 & 84.0 & 22.0 &
37.0 & 18.9 & 46.0 & 9.0 \\
\midrule
\multirow{3}{*}{HarnessVLN}
& \texttt{GPT-5.6-luna} &
76.0 & 40.8 & 86.0 & 10.0 &
55.0 & 33.0 & 74.0 & 19.0 \\
& \texttt{GPT-6-astra} &
86.0 & 47.1 & 95.0 & 9.0 &
63.0 & 39.4 & 76.0 & 13.0 \\
& \texttt{Qwen3.8-flash} &
72.0 & 36.9 & 85.0 & 13.0 &
51.0 & 32.8 & 75.0 & 24.0 \\
\bottomrule
\end{tabular}
\end{table}

\section{Unified Model Prompts}
\label{app:prompts}

The same MLLM performs initialization, planning, grounding, verification, and recovery through function-specific prompts. Each call receives the relevant task context, labeled observations, TODO progress, retrieved ST Graph evidence, and tool feedback. The following excerpts are condensed for presentation; the JSON examples show selected response fields.

\subsection{Prompt Excerpts}
\label{app:prompt_excerpts}

\begin{appPrompt}{Navigation planning: update progress before acting}
\textbf{Inputs.} [instruction], [history], [current directional views], [TODO list], [retrieved memory], [feedback], [available actions].

\smallskip
\textbf{Progress update.} Review all TODO items by their stable \texttt{todo\_id}, rather than list position. Keep at most one item active. Mark an item completed only when supported by a concrete observed result and its evidence; provide a reason for any blocked item. Revisit earlier items when later subgoals have already been completed.

\smallskip
\textbf{Action selection.} Update progress first, then select one offered navigation or backtracking action from the updated context. Set \texttt{stop} to \texttt{true} when the instruction endpoint has been reached; pending TODOs do not prevent a stop request.

\begin{lstlisting}[style=appJSON]
{
  "todo_updates": [
    {
      "todo_id": "todo-1",
      "status": "completed",
      "result": "<observed evidence of completion>",
      "evidence_ids": ["ob-000001"]
    }
  ],
  "action": "navigate to left",
  "stop": false
}
\end{lstlisting}
\end{appPrompt}

\begin{appPrompt}{ObjectNav stop verification: identify and localize}
\textbf{Inputs.} [target category], [current directional views].

\smallskip
\textbf{Verification.} Confirm that the target is clearly visible and belongs to the requested category; reject look-alikes. Do not estimate distance from RGB; the depth tool measures it.

\smallskip
\textbf{Localization.} If confident, select the clearest view and return a tight box around the object itself on a normalized $[0,1000]$ scale. Otherwise, return \texttt{CONTINUE}, a \texttt{null} view, and an empty box.

\begin{lstlisting}[style=appJSON]
{
  "analysis": "<visible evidence for the category>",
  "decision": "STOP",
  "target_view": "forward",
  "bbox_2d": [350, 200, 600, 800]
}
\end{lstlisting}
\end{appPrompt}

\subsection{Execution Contract}
\label{app:execution_contract}
A model's stop decision is a request to the Harness, which applies the geometric checks in Appendix~\ref{app:simulation} before issuing the simulator \texttt{STOP} action. Route verification checks the instruction endpoint, whereas ObjectNav verification checks the target category. Boxes use $[x_1,y_1,x_2,y_2]$. For a $W\times H$ image, normalized coordinates map to pixels as $x_{\mathrm{px}}=Wx/1000$ and $y_{\mathrm{px}}=Hy/1000$.

\section{Simulation Visualizations}
\label{app:visualizations}

We illustrate six recorded rollouts across the four benchmarks, using
\texttt{GPT-5.5} for R2R/RxR and \texttt{GPT-5.6-luna} for ObjectNav, with
one shared model across all reasoning functions in each rollout. These selected cases show execution behavior and failure modes; they are not aggregate performance estimates.

\subsection{Runtime Dashboard and Interpretation}
\label{app:dashboard}

\begin{figure}[htbp]
\centering
\includegraphics[width=\linewidth]{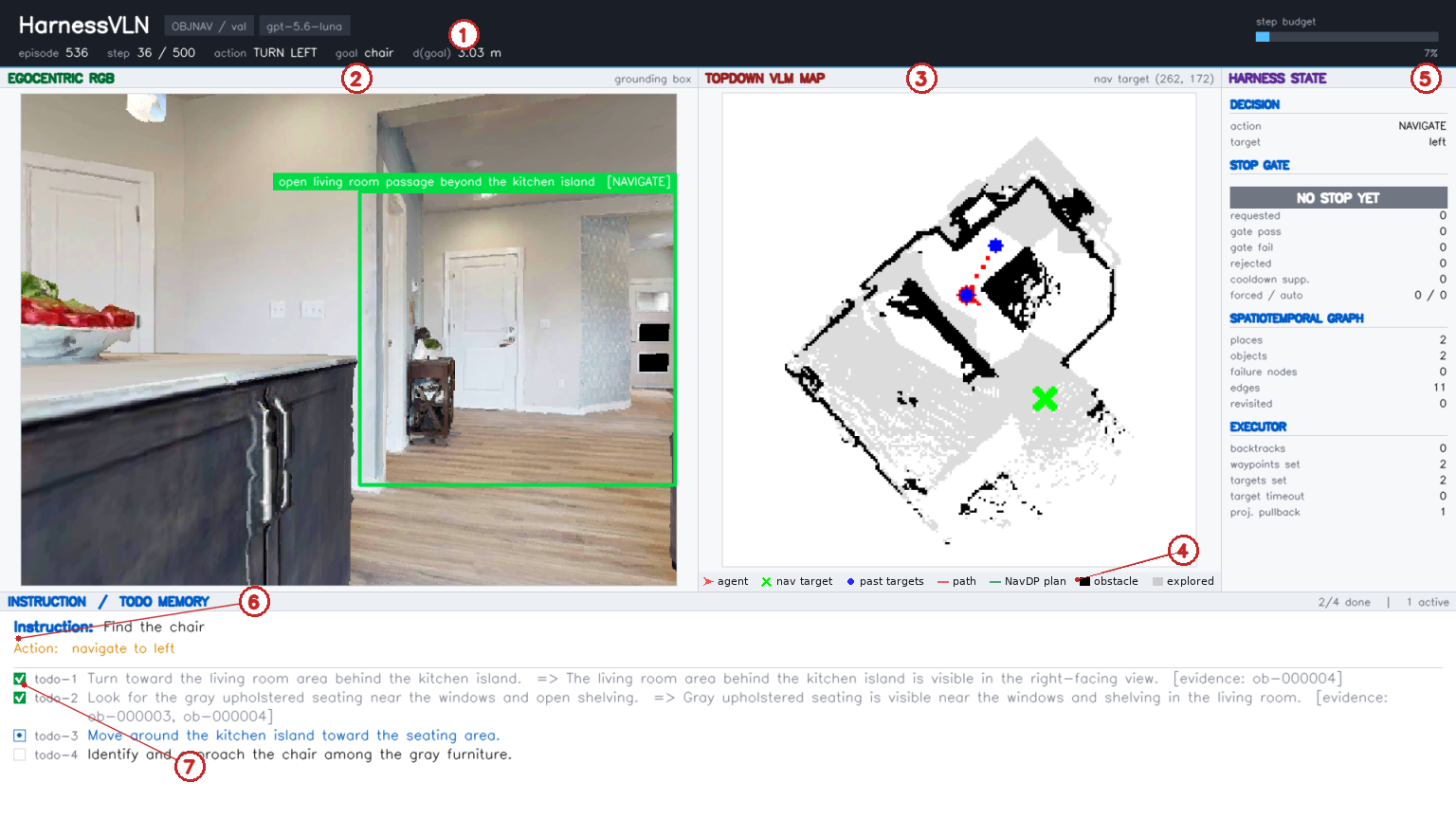}
\caption{\textbf{Runtime dashboard.} HM3D-v2 episode 536 at step 36.
The numbered panels connect the observation, navigation state, and progress memory; the key below describes their contents.}
\label{fig:app_dashboard}
\smallskip
{\small
\begin{tabularx}{\linewidth}{@{}rX@{}}
\textbf{1} & \textbf{Run status:} model, episode, step budget, action, and evaluator goal distance. \\
\textbf{2} & \textbf{RGB observation:} the current view and any grounding box recorded at that step. \\
\textbf{3} & \textbf{Navigation map:} occupancy, trajectory, and projected navigation targets. \\
\textbf{4} & \textbf{Map key:} agent, current/past targets, path, plan, obstacles, and explored area. \\
\textbf{5} & \textbf{Harness state:} decision, stop checks, ST Graph counts, and executor counters. \\
\textbf{6} & \textbf{Task and action:} the instruction or target category and current navigation choice. \\
\textbf{7} & \textbf{TODO memory:} subgoals, status updates, and supporting observations. \\
\end{tabularx}
}
\end{figure}

The map is a geometric navigation map; the ST Graph is summarized by counts in panel 5. The header's goal distance is an evaluator diagnostic, not an agent input. In the case figures, each row contains two recorded RGB views and the final runtime map. Frame indices and stop-check event indices are reported separately; a verification view can differ from the displayed forward RGB view. Captions report stop-check distances; terminal benchmark distances are explicitly labeled as goal distances.

\clearpage
\subsection{Successful Navigation across Tasks}
\label{app:successful_cases}

\begin{figure}[htbp]
\centering
\begin{tabularx}{\linewidth}{@{}*{3}{>{\centering\arraybackslash}X}@{}}
RGB: step 51 & RGB: step 150 & Map: step 150
\end{tabularx}
\includegraphics[width=\linewidth]{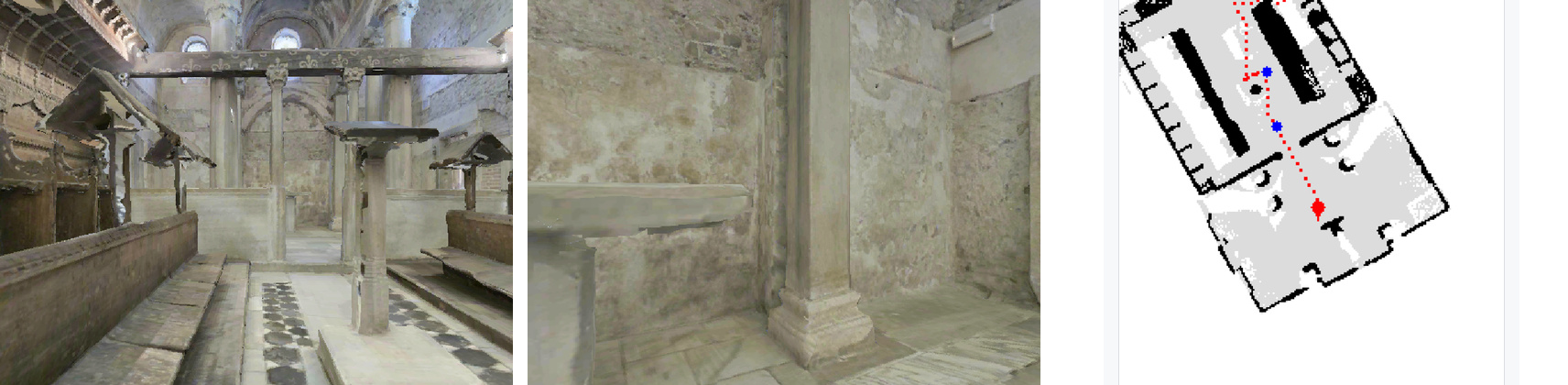}
\caption{\textbf{R2R 650: reaching a route endpoint.}
The instruction asks the agent to cross the room and wait in the archway in front of the podium. The projected-target check passes at step 149 (0.56\,m); the episode succeeds at step 150 with a terminal goal distance of 2.39\,m.}
\label{fig:app_r2r650}
\end{figure}

\begin{figure}[htbp]
\centering
\begin{tabularx}{\linewidth}{@{}*{3}{>{\centering\arraybackslash}X}@{}}
RGB: step 0 & RGB: step 926 & Map: step 926
\end{tabularx}
\includegraphics[width=\linewidth]{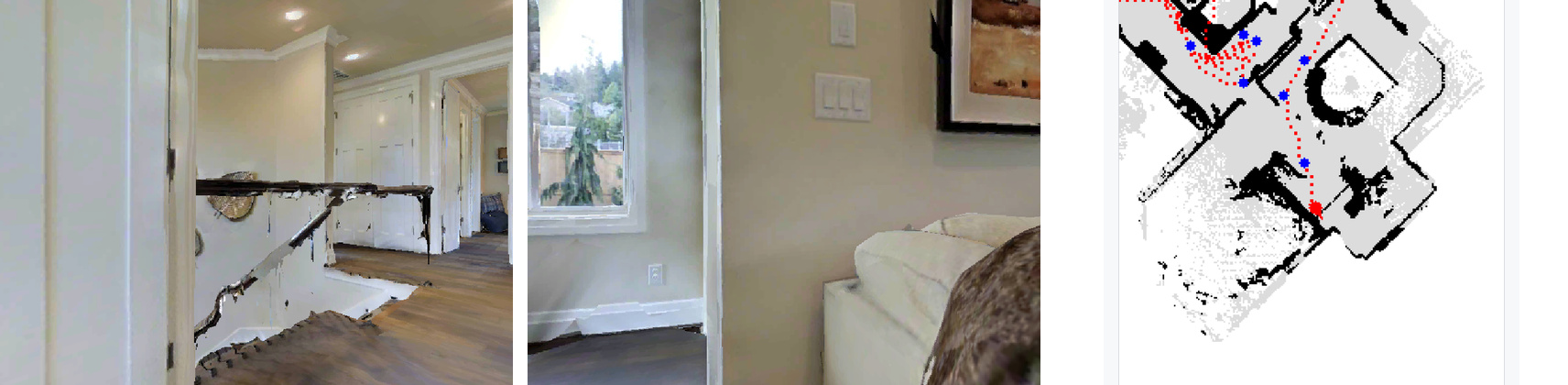}
\caption{\textbf{RxR 154: a long instruction-following rollout under the extended step budget.}
The agent navigates from the stairs through the dining and kitchen areas to the endpoint near a glass door. The projected-target check passes at step 925 (0.49\,m), and the episode succeeds at step 926 with a terminal goal distance of 0.43\,m. This rollout is from the 1,000-step diagnostic setting; main evaluations use the default 500-step budget.}
\label{fig:app_rxr154}
\end{figure}

\begin{figure}[htbp]
\centering
\begin{tabularx}{\linewidth}{@{}*{3}{>{\centering\arraybackslash}X}@{}}
RGB: step 36 & RGB: step 95 & Map: step 95
\end{tabularx}
\includegraphics[width=\linewidth]{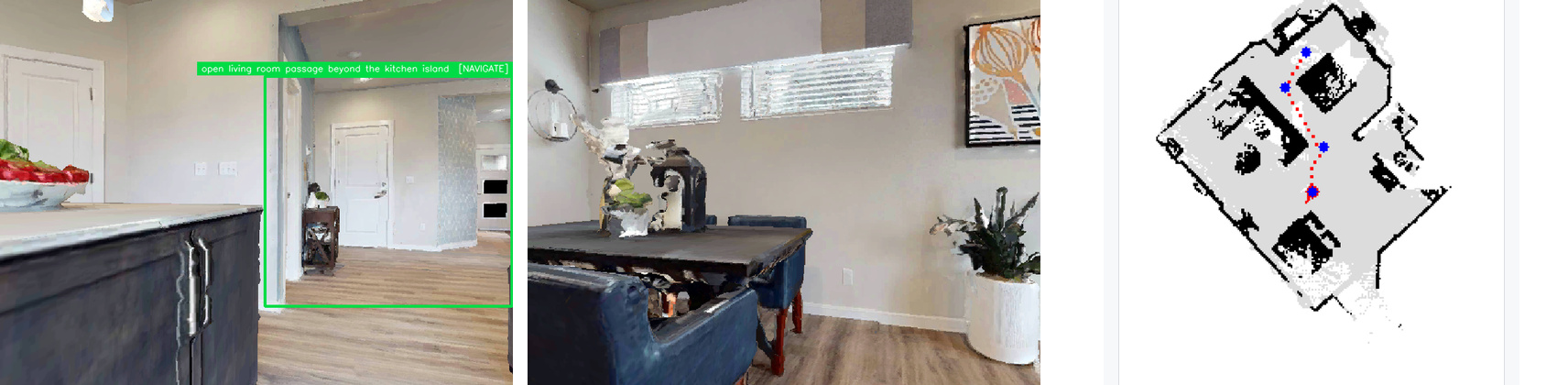}
\caption{\textbf{HM3D-v2 536: finding a chair.}
The early view grounds a passage into the seating area; the later view shows the final position. A fresh-depth check passes at step 94 (0.64\,m), followed by success at step 95 with a terminal goal distance of 0.05\,m.}
\label{fig:app_hm3d536}
\end{figure}

\clearpage
\subsection{Stop Verification and Failure Cases}
\label{app:failure_cases}

\begin{figure}[htbp]
\centering
\begin{tabularx}{\linewidth}{@{}*{3}{>{\centering\arraybackslash}X}@{}}
RGB: step 216 & RGB: step 430 & Map: step 430
\end{tabularx}
\includegraphics[width=\linewidth]{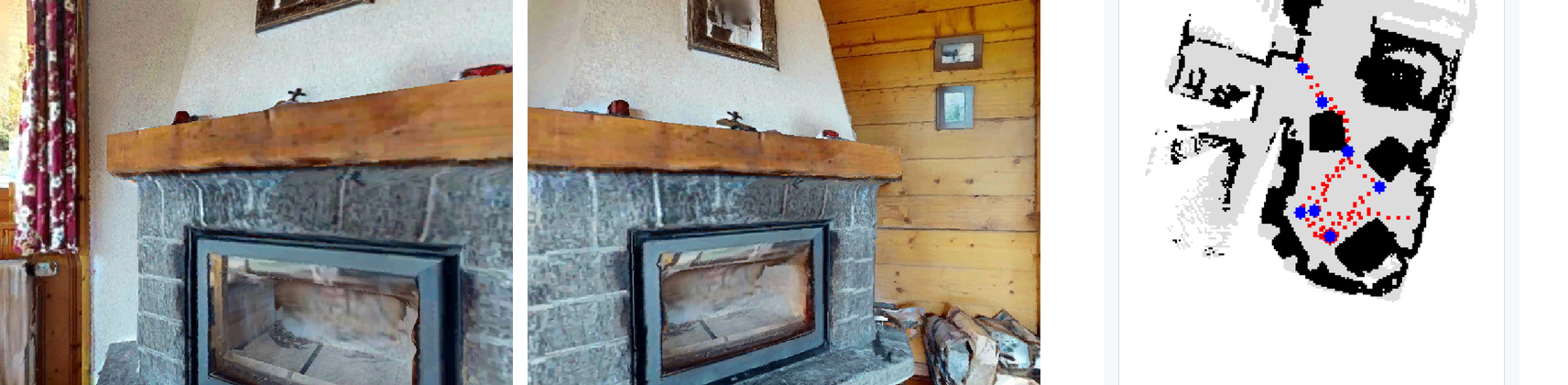}
\caption{\textbf{HM3D-OVON 1297: a rejected stop followed by success.}
While searching for a picture, fresh depth rejects stopping at step 216 (4.65\,m), despite a projected-target distance of 0.40\,m. A later depth check passes at step 429 (1.53\,m); the episode succeeds at step 430 with a terminal goal distance of 0.13\,m.}
\label{fig:app_ovon1297}
\end{figure}

\begin{figure}[htbp]
\centering
\begin{tabularx}{\linewidth}{@{}*{3}{>{\centering\arraybackslash}X}@{}}
RGB: step 178 & RGB: step 475 & Map: step 475
\end{tabularx}
\includegraphics[width=\linewidth]{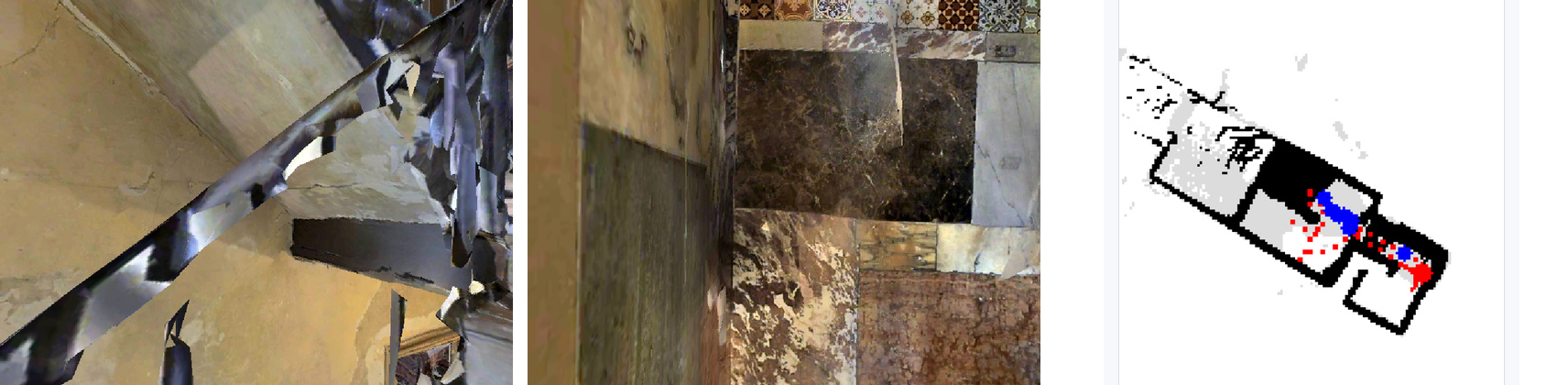}
\caption{\textbf{R2R 839: failing to reach the instructed room.}
The task asks the agent to descend the stairs and enter the room on the left. The projected-target check eventually passes at step 474 (0.998\,m), but termination is 6.48\,m from the goal. Proximity to the selected target does not establish that the instruction endpoint is correct.}
\label{fig:app_r2r839}
\end{figure}

\begin{figure}[htbp]
\centering
\begin{tabularx}{\linewidth}{@{}*{3}{>{\centering\arraybackslash}X}@{}}
RGB: step 16 & RGB: step 47 & Map: step 47
\end{tabularx}
\includegraphics[width=\linewidth]{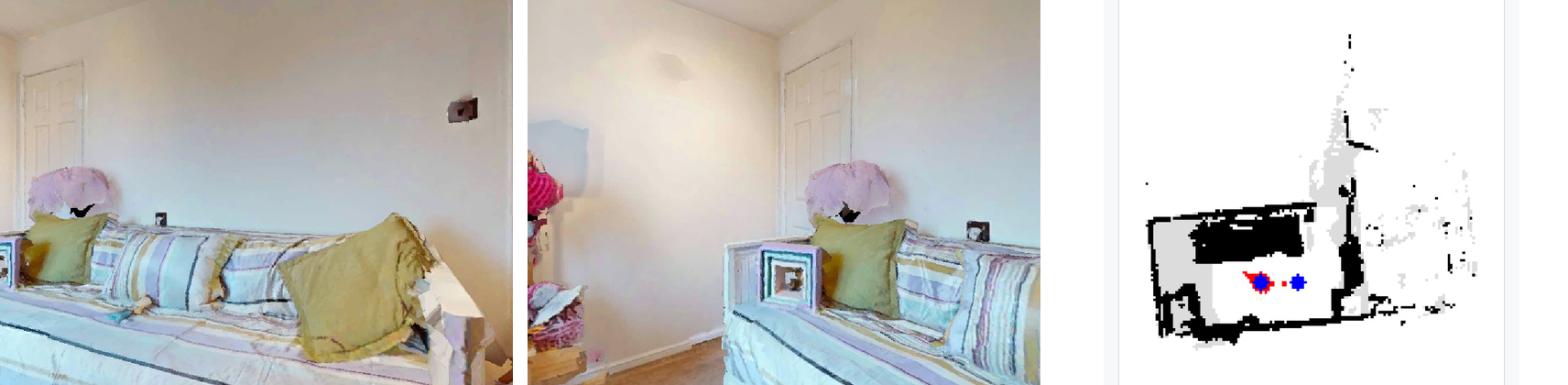}
\caption{\textbf{HM3D-OVON 2469: passing the gate outside the success region.}
For the target category \emph{picture}, fresh depth passes at step 46 (1.20\,m). The episode stops at step 47 with a benchmark goal distance of 2.15\,m and fails. Sensor proximity and benchmark success remain distinct.}
\label{fig:app_ovon2469}
\end{figure}

\end{document}